\documentclass[sn-mathphys-num]{sn-jnl}

\usepackage{graphicx}%
\usepackage{multirow}%
\usepackage{amsmath,amssymb,amsfonts}%
\usepackage{amsthm}%
\usepackage{mathrsfs}%
\usepackage[title]{appendix}%
\usepackage{textcomp}%
\usepackage{manyfoot}%
\usepackage{booktabs}%
\usepackage{algorithm}%
\usepackage{algorithmicx}%
\usepackage{algpseudocode}%
\usepackage{listings}%
\usepackage{comment}
\usepackage{subfigure}
\newcommand{\etal}{\emph{et al}.}
\newcommand{\eg}{\emph{e.g}.}
\newcommand{\ie}{\emph{i.e}.}
\newcommand{\etc}{\emph{etc}.} \newcommand{\vs}{\emph{vs}.}

\usepackage{colortbl}
\usepackage{color}
\usepackage[dvipsnames]{xcolor}
\usepackage{cite}
\usepackage{url}
\usepackage{indentfirst} 
\usepackage{natbib}
\usepackage{dsfont}
\usepackage{pifont}
\usepackage{epsfig}
\usepackage{epstopdf}
\usepackage{makecell,multirow,diagbox}
\usepackage{soul}
\usepackage[utf8]{inputenc}
\usepackage{mathtools,xspace}
\usepackage{ragged2e}

\theoremstyle{thmstyleone}%
\theoremstyle{thmstyletwo}%

\theoremstyle{thmstylethree}%

\begin{document}

\title{DeepSolo++: Let Transformer Decoder with Explicit Points Solo for Multilingual Text Spotting}


\author[1]{\fnm{Maoyuan} \sur{Ye}}\email{yemaoyuan@whu.edu.cn}
\equalcont{These authors contributed equally to this work.}

\author[2]{\fnm{Jing} \sur{Zhang}}\email{jing.zhang1@sydney.edu.au}
\equalcont{These authors contributed equally to this work.}

\author[3]{\fnm{Shanshan} \sur{Zhao}}\email{sshan.zhao00@gmail.com}

\author*[1]{\fnm{Juhua} \sur{Liu}}\email{liujuhua@whu.edu.cn}

\author[2]{\fnm{Tongliang} \sur{Liu}}\email{tongliang.liu@sydney.edu.au}

\author*[1]{\fnm{Bo} \sur{Du}}\email{dubo@whu.edu.cn}

\author[4]{\fnm{Dacheng} \sur{Tao}}\email{dacheng.tao@ntu.edu.sg}

\affil[1]{School of Computer Science, National Engineering Research Center for Multimedia Software, Institute of Artificial Intelligence, and Hubei Key Laboratory of Multimedia and Network Communication Engineering, Wuhan University, Wuhan, China}

\affil[2]{School of Computer Science, Faculty of Engineering, The University of Sydney, Australia}

\affil[3]{JD Explore Academy at JD.com, China}

\affil[4]{College of Computing \& Data Science at Nanyang Technological University, Singapore}


\abstract{
End-to-end text spotting aims to integrate scene text detection and recognition into a unified framework. Dealing with the relationship between the two sub-tasks plays a pivotal role in designing effective spotters. Although Transformer-based methods eliminate the heuristic post-processing, they still suffer from the synergy issue between the sub-tasks and low training efficiency. Besides, they overlook the exploring on multilingual text spotting which requires an extra script identification task. In this paper, we present \textbf{DeepSolo++}, a simple DETR-like baseline that lets a single \textbf{De}coder with \textbf{E}xplicit \textbf{P}oints \textbf{Solo} for text detection, recognition, and script identification simultaneously. Technically, for each text instance, we represent the character sequence as ordered points and model them with learnable explicit point queries. After passing a single decoder, the point queries have encoded requisite text semantics and locations, thus can be further decoded to the center line, boundary, script, and confidence of text via very simple prediction heads in parallel. Furthermore, we show the surprisingly good extensibility of our method, in terms of character class, language type, and task. On the one hand, our method not only performs well in English scenes but also masters the transcription with complex font structure and a thousand-level character classes, such as Chinese. On the other hand, our DeepSolo++ achieves better performance on the additionally introduced script identification task with a simpler training pipeline compared with previous methods. Extensive experiments on public benchmarks demonstrate that our simple approach achieves better training efficiency compared with Transformer-based models and outperforms the previous state-of-the-art. For example, on ICDAR 2019 ReCTS for Chinese text, our method boosts the 1-NED metric to a new record of 78.3\%. On ICDAR 2019 MLT, DeepSolo++ achieves absolute 5.5\% H-mean and 8.0\% AP improvements on joint detection and script identification task, and 2.7\% H-mean gains on end-to-end spotting. In addition, our models are also compatible with line annotations, which require much less annotation cost than polygons. The code is available at \href{https://github.com/ViTAE-Transformer/DeepSolo}{DeepSolo}.
}




\maketitle

\section{Introduction}
\label{sec:introduction}
Detecting and recognizing text in natural scenes, \textit{a.k.a.} text spotting, has drawn increasing attention due to its wide range of applications \citep{zhang2020empowering,long2021scene}, such as intelligent navigation \citep{desouza2002vision}. 
How to deal with the relationship between detection and recognition is a long-standing and open problem, which has a significant impact on the structural pipeline, performance, efficiency, annotation cost, \etc

\begin{figure*}[!t]
    \centering
    \includegraphics[width=\linewidth]{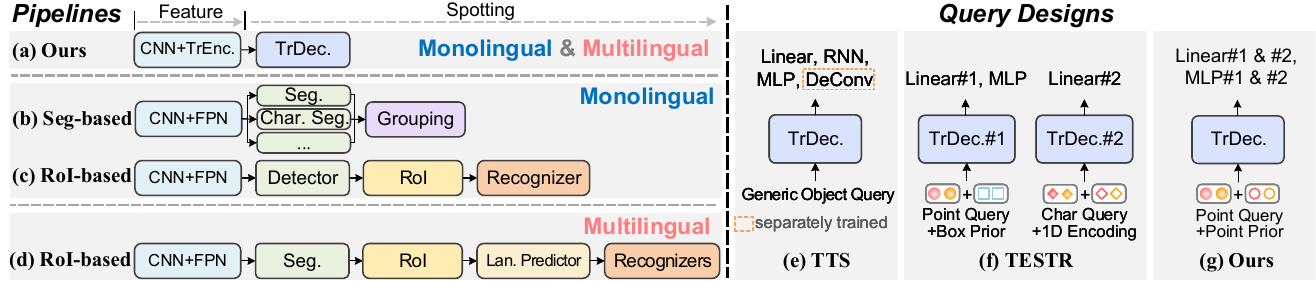}
    \caption{Comparison of text spotting pipelines and query designs. In our method, for both monolingual and multilingual text spotting, the spotting part is a solo by the Transformer decoder with explicit points. `TrEnc.' (`TrDec.'): Transformer encoder (decoder). `Char.': characters. `Seg.': segmentation. `Lan. Predictor': language prediction network.}
    \label{fig:1}
\end{figure*}

Most pioneering end-to-end spotting methods focus on one specific language scene, especially English, without launching a unified model for several languages. The majority of them \citep{liu2020abcnet,feng2021residual,wang2021pan++,liao2020mask,ronen2022glass} follow a detect-then-recognize pipeline, which first detects text instances and then exploits crafted Region-of-Interest (RoI) operation to extract features within the detected area, finally feeds them into a following recognizer (Fig.~\ref{fig:1}(c)). Although these methods have achieved great progress, there are two main limitations. 1) The extra RoI transform for feature alignment is indispensable. Some RoI operations require polygon annotations, which are not applicable when only weak annotations (\eg, lines) are available. 2) Additional efforts are desired to address the synergy issue between the detection and recognition modules \citep{zhong2021arts, huang2022swintextspotter}. 
In contrast, segmentation-based methods \citep{xing2019convolutional,wang2021pgnet} try to isolate the two sub-tasks and conduct spotting in a parallel multi-task framework with a shared backbone (Fig.~\ref{fig:1}(b)).
Nevertheless, they require grouping post-processing to gather unstructured components.

When it comes to multilingual text spotting, existing methods \citep{buvsta2019e2e,baek2020character,Huang_2021_CVPR,huang2023task} still lie in the detect-then-recognize flow. In addition, to solve the further introduced script identification task and allow for more convenient customization of recognizers for different languages, Huang \etal ~\citep{Huang_2021_CVPR,huang2023task} devise tailored language predictors and use them for routing to a recognizer. Despite the novel framework, the overall pipeline undergoes unfortunate expansion, growing in complexity with the added modules (Fig.~\ref{fig:1}(d)). Extra efforts are involved in designing a language predictor and exploring a module-wise multi-stage training pipeline.

In recent years, Transformer \citep{vaswani2017attention} has shown eminent flexibility and sparked the performance remarkably for various computer vision tasks \citep{dosovitskiy2020image,liu2021swin,xu2021vitae}, including text spotting. Although the text spotters \citep{zhang2022text,kittenplon2022towards} based on DETR \citep{carion2020end} can get rid of RoI and heuristic post-processing, they lack efficient and joint representation to deal with scene text detection and recognition, \eg, requiring an extra RNN module in TTS \citep{kittenplon2022towards} (Fig.~\ref{fig:1}(e)) or exploiting individual Transformer decoder for each sub-task in TESTR \citep{zhang2022text} (Fig.~\ref{fig:1}(f)). The generic object query exploited in TTS fails to consider the unique characteristics of scene text, \eg, location and shape. While TESTR adopts point queries with box positional prior that is coarse for point predicting \citep{ye2022dptext}, and the queries are different between detection and recognition, introducing unexpected heterogeneity. Consequently, these designs hamper the performance and training efficiency. Last but not least, the potential of Transformer on multilingual text spotting is unexplored.

In this paper, we propose a novel query form based on explicit point representations of text lines. Built upon it, we present a simple DETR-like baseline that lets a single \textbf{De}coder with \textbf{E}xplicit \textbf{P}oints \textbf{Solo} (dubbed \textbf{DeepSolo++}) for multilingual text detection, recognition, and script identification all at once (Fig.~\ref{fig:1}(a) and Fig. \ref{fig:1}(g)). Technically, for each instance, we first represent the character sequence as ordered points, where each point has explicit attributes of position, offsets to the top and bottom boundary, and category. Specifically, we devise top-$K$ Bezier center curves to fit scene text instances with arbitrary shapes and sample a fixed number of on-curve points. Then, we leverage the sampled points to generate positional queries and guide the learnable content queries with explicit positional prior. After flowing out the decoder, the output queries are expected to have encoded requisite text semantics and locations. Finally, we adopt several simple prediction heads (a linear layer or MLP) in parallel to decode the queries into the center line, boundary, transcript, and confidence of text, thereby solving detection and recognition simultaneously. For multilingual spotting, we introduce an extra script token with explicit point information and use it for the script identification task and routing to a specific linear layer for character classification. We also devise a script-aware bipartite matching scheme for training DeepSolo++. Note that we denote the monolingual text spotter as DeepSolo, without the script token and corresponding designs for multilingual tasks.

In summary, the main contributions are four-fold: 
\begin{itemize}
    \item We propose a novel query form based on explicit points sampled from the Bezier center curve representation of text instance lines, which can efficiently encode the position, shape, and semantics of text, thus helping simplify both the monolingual and multilingual text spotting pipeline.
    \item We propose simple yet effective baseline models named DeepSolo and DeepSolo++ for monolingual and multilingual text spotting, respectively.
    \item Our method shows several good properties, including 1) \textbf{simplicity} of structure and training pipeline, 2) \textbf{efficiency} of training and inference, and 3) \textbf{extensibility} of character class, language, and task.
    \item Extensive experiments on challenging datasets demonstrate the state-of-the-art (SOTA) performance of our method and some other distinctions, such as the effectiveness on dense and long text, and the flexibility of position annotation form.
\end{itemize}

The rest of this paper is organized as follows. In Sec.~\ref{sec:relatedworks}, we briefly review the related works. In Sec.~\ref{sec:method}, we introduce our methodology in detail. Sec.~\ref{sec:exp} report our experimental results. A discussion is presented in Sec.~\ref{sec:discussion}. Finally, we conclude our study in Sec.~\ref{sec:conclusion}.

\section{Related Work}
\label{sec:relatedworks}
\subsection{RoI-based Text Spotters}
Taking the merit of joint optimization of detection and recognition, recent works dive into developing end-to-end text spotters. Most of them craft RoI \citep{he2017mask} or Thin-Plate Splines (TPS) \citep{bookstein1989principal} to bridge the detector and recognizer. 
Mask TextSpotter series \citep{lyu2018mask, liao2020mask, liao2021mask} conduct character segmentation for recognition based on the RoIAligned features. GLASS \citep{ronen2022glass} devises a plug-in global-to-local attention module to enhance the representation ability with the assistance of Rotated-RoIAlign. 
To better rectify curved texts, ABCNet \citep{liu2020abcnet,liu2021abcnet} proposes the BezierAlign module using a parameterized Bezier curve. However, they ignore the synergy issue \citep{zhong2021arts, huang2022swintextspotter} between the two tasks. 
To overcome this dilemma, SwinTextSpotter \citep{huang2022swintextspotter} proposes a Recognition Conversion module to back-propagate recognition information to the detector. Although the above methods have achieved remarkable progress, they require an extra RoI-based or TPS-based module. Since some RoI operations require polygon annotations, the methods may not apply to scenarios with only weak position annotations (\eg, single points, lines). Moreover, a more effective and simpler solution is also expected to address the synergy issue.

\subsection{RoI-free Text Spotters}
Inspired by DETR family \citep{carion2020end, zhu2020deformable}, recent works \citep{kittenplon2022towards,zhang2022text} explore the Transformer framework without RoIAlign and complicated post-processing. 
TTS \citep{kittenplon2022towards} adds an RNN recognizer into Deformable-DETR \citep{zhu2020deformable} and shows its potential on weak annotations. TESTR \citep{zhang2022text} adopts two parallel Transformer decoders for detection and recognition. Although these methods unlock the potential of Transformer in text spotting, there are still some limitations. For example, the vanilla or individual queries used in TTS and TESTR cannot efficiently encode text features (\eg, location, shape, and semantics), affecting the training efficiency \citep{ye2022dptext} and even increasing the model complexity. Besides, the axis-aligned box annotations used in TTS might not be ideal enough \citep{peng2022spts} for scene text since the box contains a certain portion of background regions and even other texts, thus introducing extra noise. In our work, thanks to the proposed novel explicit query form, text detection and recognition tasks enjoy an efficient and unified pivot. Moreover, DeepSolo is also compatible with weak text position annotation form, \ie, text center line, which is less affected by background and other instances. 

\subsection{Multilingual Text Spotting}
The majority of existing works focus on training one model using data from a specific language. However, they did not unveil the capability of identifying script and recognizing text instances varying from multiple languages in a single unified network. Only a few works investigated end-to-end multilingual text spotters and achieved exciting results. Among them, E2E-MLT \citep{buvsta2019e2e} proposes a framework for multilingual text spotting by handling characters from all languages in the same recognizer, the same as CRAFTS \citep{baek2020character}. In comparison, Multiplexed TextSpotter \citep{Huang_2021_CVPR} and Grouped TextSpotter \citep{huang2023task} leverage several adapted recognizers to deal with different languages respectively, with Language Prediction Network (LPN) routing the detected text to an appropriate recognizer. However, all of the above methods rely on RoI and sequence-to-sequence recognizers. Besides, in \citep{Huang_2021_CVPR,huang2023task}, hand-crafted LPNs are required for script identification and multi-head routing. From a different paradigm, we demonstrate that our DeepSolo++ with a simple routing scheme for multilingual recognition, can reach higher performance without a RoI operation, neither with a specific language prediction network nor adapted recognizers.

\subsection{Comparison to the Conference Version}
A preliminary version of this work was presented in \citep{ye2022deepsolo}, where a monolingual text spotter (DeepSolo) was introduced. 
Our current study expands significantly, incorporating three major improvements.
\begin{enumerate}
    \item We investigate the performance of DeepSolo on Chinese scene text, characterized by more complex font structures and a larger character set. DeepSolo delivers SOTA 1-NED result on ICDAR 2019 ReCTS \citep{zhang2019icdar}.
    It demonstrates the inherent simplicity and remarkable extensibility of DeepSolo in tackling challenging language scenes.
    \item We propose a new baseline model dubbed DeepSolo++ for multilingual text spotting which involves an extra script identification task. DeepSolo++ simplifies the multilingual text spotting pipeline (Fig.~\ref{fig:1}(a) \vs ~Fig.~\ref{fig:1}(d)) and achieves better performance with a simpler training pipeline on the benchmarks \citep{nayef2017icdar2017,nayef2019icdar2019}. It validates the extensibility of our method on a new task.
    \item Additional experiment results, encompassing the effectiveness of DeepSolo on dense and long scene text, its robustness in handling highly rotated instances, and other relevant factors, are presented. Moreover, comprehensive ablation studies, in-depth analyses, and supplementary visualization results are provided to augment the research findings.
\end{enumerate}

\section{Methodology}
\label{sec:method}
We propose the \textbf{DeepSolo++} for multilingual text spotting, which retains structural simplicity by leveraging the single decoder to solo for multilingual text detection, recognition, and script identification. We denote the monolingual text spotting part as DeepSolo, without script tokens introduced in Sec.~\ref{subsec:point_query} and other corresponding designs for multilingual tasks.

\subsection{Overview}
\label{sec:overview}
\begin{figure*}[!t]
    \centering
    \includegraphics[width=1\linewidth]{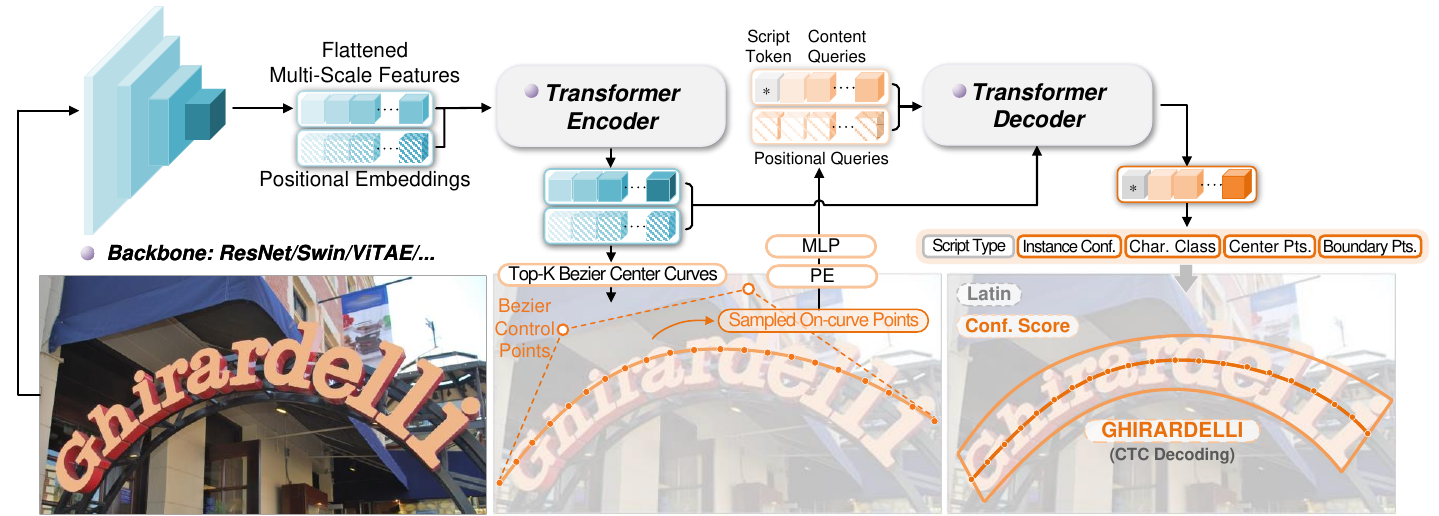}
    \caption{The architecture of DeepSolo++. We propose an explicit query form based on the points sampled from the Bezier center curve representation of text, solving multilingual text spotting with a single decoder and simple prediction heads in a concise framework.}
    \label{fig:model}
\end{figure*}

\noindent\textbf{Preliminary.} Bezier curve, firstly introduced into text spotting by ABCNet \citep{liu2020abcnet}, can flexibly fit the arbitrarily shaped scene text with minimal control points. Different from ABCNet which crops features with BezierAlign, we explore the distinctive utilization of the Bezier curve in the DETR framework. More details of the Bezier curve generation can be referred to \citep{liu2020abcnet, liu2021abcnet}. Given four Bezier control points \citep{liu2020abcnet} 
for each side (top and bottom) of a text instance, we simply compute Bezier control points for the center curve by averaging the corresponding control points on the top and bottom sides. Then, $N$ points are uniformly sampled on the center, top, and bottom curves, respectively, serving as the ground truth. Note that the order of on-curve points should be in line with the text reading order.

\noindent\textbf{Model Architecture}. The overall architecture of DeepSolo++ is depicted in Fig.~\ref{fig:model}. After receiving the image features from the encoder, the Bezier center curve proposal represented by four Bezier control points and the corresponding score are generated. Then, the top-$K$ scored proposals are selected. For each selected proposal, a certain number of points are uniformly sampled on the curve. These point coordinates are encoded as positional queries and added to learnable content queries, forming composite queries. Next, the composite queries are fed into the decoder to gather useful features via deformable cross-attention \citep{zhu2020deformable}. Following the decoder, several simple prediction heads are adopted, each responsible for solving a specific task. 

\subsection{Top-$K$ Bezier Center Curve Proposals}
\label{sec:top-k proposal}
Different from the box proposal adopted in previous works \citep{zhu2020deformable, zhang2022text, ye2022dptext}, which has the drawbacks in representing text with arbitrary shape, we design a simple Bezier center curve proposal scheme from the text center line perspective. It can efficiently fit arbitrarily shaped scene text with minimal control points and distinguish one from others, making using line annotations possible. Specifically, given the image features, on each pixel of the feature maps, a 3-layer MLP (8-dim in the last layer) is used to predict offsets to four Bezier control points, determining a curve that represents one text instance. Let $i$ index a pixel from features at level $l \in \left\{1,2,3,4\right\}$ with 2D normalized coordinates $\hat{p}_i = \left(\hat{p}_{ix}, \hat{p}_{iy}\right) \in \left[0,1\right]^2$, its corresponding Bezier control points $BP_i = \{\bar{bp}_{i_0}, \bar{bp}_{i_1}, \bar{bp}_{i_2}, \bar{bp}_{i_3}\}$ are predicted:
\begin{equation}
    \bar{bp}_{i_j} = (\sigma(\Delta p_{ix_j} + \sigma^{-1}(\hat{p}_{ix})), \sigma(\Delta p_{iy_j} + \sigma^{-1}(\hat{p}_{iy}))), \label{eq_1}
\end{equation}
where $j \in \{0,1,2,3\}$, $\sigma$ is the sigmoid function. MLP head only predicts the offsets $\Delta p_{i\{x_0,y_0,\ldots,x_3,y_3\}}$. Moreover, we use a linear layer for text or non-text classification. And top-$K$ scored curves are selected as proposals.

\subsection{Explicit Point Query}
\label{subsec:point_query}
\noindent \textbf{Query Initialization.} Given the top-$K$ proposals $\hat{BP}_k$ ($k \in \{0,1,\ldots,K-1\}$) selected from $BP_i$, we uniformly sample $N$ points on each curve according to the Bernstein Polynomials \citep{lorentz2013bernstein} which is implemented by a simple matrix product in practice. Here, we get normalized point coordinates $Coords$ of shape $K \times N \times 2$ in each image. The point positional queries $\textbf{P}_{q}$ of shape $K \times N \times 256$ are generated by:
\begin{equation}
    \textbf{P}_{q} = MLP(PE(Coords)), \label{eq_2}
\end{equation}
where $PE$ represents the sinusoidal positional encoding function. Following Liu \etal ~\citep{liu2022dab}, we also adopt a 2-layer $MLP$ head with $ReLU$ activation for further projection. On the other side, we initialize point content queries $\textbf{C}_{q}$ using learnable embeddings. Then, we add $\textbf{P}_{q}$ to $\textbf{C}_{q}$ to get the composite queries $\textbf{Q}_q$:
\begin{equation}
    \textbf{Q}_q = \textbf{C}_{q} + \textbf{P}_{q}. \label{eq_3}
\end{equation}
We empirically exploit unshared point embeddings for $\textbf{C}_{q}$, \ie, $N$ point content queries of shape $N \times 256$ in one text instance are used for each of the $K$ proposals.

\noindent \textbf{Query Update in the Decoder.} After obtaining the composite queries $\textbf{Q}_q$, we feed them into the Transformer decoder. We follow previous works \citep{zhang2022text,ye2022dptext,du2022i3cl} to firstly mine the relationship between queries within one text instance using an intra-group self-attention across dimension $N$. Here, keys are the same with queries while values only contain the content part: $\textbf{K}_q = \textbf{Q}_q$, $\textbf{V}_q = \textbf{C}_q$. Then, an inter-group self-attention is conducted across $K$ instances to capture the relationship between different instances. The updated composite queries are further sent into the deformable cross-attention to aggregate multi-scale text features from the encoder. The point coordinates $Coords$ are used as the reference points in the deformable attention. 
With the explicit point information flowing in the decoder, we adopt a 3-layer MLP head to predict the offsets and update point coordinates after each decoder layer. Then, the updated coordinates will be transformed into new positional queries by Eq.~(\ref{eq_2}).

\begin{figure}[!t]
    \centering
    \includegraphics[width=0.85\linewidth]{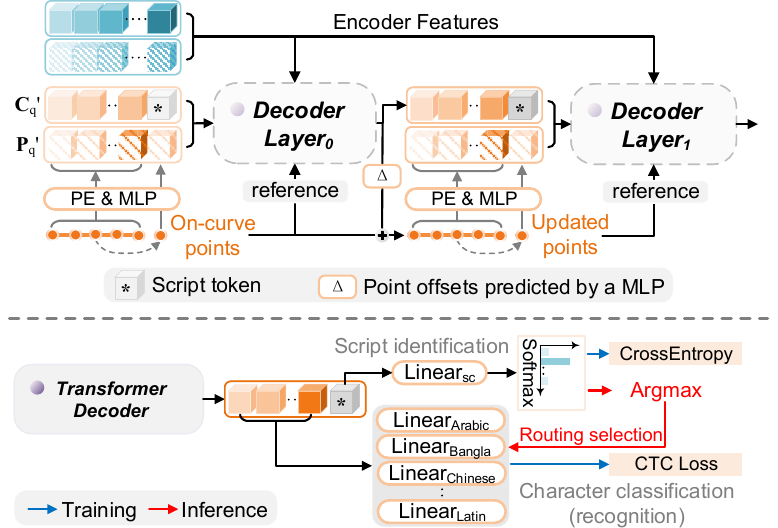}
    \caption{The illustration of query modeling (top) in DeepSolo++ and the pipeline of training and inference (bottom). For ease of illustration, only the queries for one text instance are plotted, and only the linear layers for script identification and character classification are shown in the bottom section.}
    \label{fig:deepsolo++}
\end{figure}

\noindent \textbf{Script Token Modeling.} 
For multilingual text spotting which involves an extra script identification task, we additionally introduce a script token with explicit point position to gather the global information of each text instance. With the script token, we can conduct script identification with a linear layer and select the routing recognition result conveniently. The illustration is presented in the top section of Fig.~\ref{fig:deepsolo++}. Specifically, for script tokens $\textbf{T}_{script}$, we select the center points of center curve proposals to generate the positional part $\textbf{P}_{center}$ referring to Eq.~(\ref{eq_2}). Then, new content queries $\textbf{C}_{q}'$ and positional queries $\textbf{P}_{q}'$ can be obtained by:
\begin{equation}
    \textbf{C}_{q}' = [\textbf{C}_{q};\textbf{T}_{script}],
    \textbf{P}_{q}' = [\textbf{P}_{q};\textbf{P}_{center}].
\end{equation}
The composite queries $\textbf{Q}_{q}'$ of shape $K \times (N+1) \times 256$ can be achieved by adding $\textbf{C}_{q}'$ and $\textbf{P}_{q}'$. While operating the intra-group self-attention across dimension $N+1$, there is an attention mask to avoid each script token seeing itself and point queries seeing the script token. Between decoder layers, after updating the points, new center points of center curves are extracted and used for generating the positional part for script tokens.

\subsection{Parallel Prediction}
After getting the queries from the decoder, we adopt simple prediction heads to solve the following sub-tasks. 
\textbf{(1) Script identification.} A linear layer $Linear_{sc}$ is employed for script identification based on the script tokens, as shown in the bottom part of Fig.~\ref{fig:deepsolo++}. During inference, the output of $Linear_{sc}$ undergoes the $softmax$ and $argmax$ operations to determine the route to a specific character classification head.
We simply train $Linear_{sc}$ and the linear layers for character classification in parallel. Note that DeepSolo for monolingual text spotting does not contain $Linear_{sc}$. 
\textbf{(2) Instance confidence score.} We use a linear projection for binary classification (text or background) with the queries of shape $K \times N \times 256$. During inference, we take the mean of $N$ scores as the confidence score for each instance.
\textbf{(3) Character classification.} As the points are uniformly sampled on the center curve of each text instance, each point query represents a specific class (including background). Following \citep{Huang_2021_CVPR,huang2023task}, we explore a multi-head routing scheme for multilingual text recognition. For multilingual text spotting in DeepSolo++, an individual linear layer takes charge in a specific script type. The setting for different scripts is shown in Tab.~\ref{tab:multilingual_linear_layer}. In DeepSolo, only a linear projection is used to perform character classification.
\textbf{(4) Center curve points.} Given the explicit point coordinates $Coords$, a 3-layer MLP head $MLP_{coord}$ is used to predict the coordinate offsets from reference points to ground truth points on the center curve.
\textbf{(5) Boundary points.} Similarly, a 3-layer MLP head $MLP_{bd}$ is used to predict the offsets to the ground truth points on the top and bottom curves.

\begin{table}[!t]
    \centering
    \caption{The setting of linear layers for different scripts.}
    \label{tab:multilingual_linear_layer}
    \begin{tabular}{c|c|c|c}
    \toprule[1.1pt]
    Linear Layer & Out Channel & Linear Layer & Out Channel \\
    \midrule[1.1pt]
    Arabic & 73 & Bangla & 110 \\
    Chinese &5,198 &Hindi &108 \\
    Japanese &2,295 &Korean &1,798 \\
    Latin &243 &Symbols &55 \\
    \bottomrule[1.1pt]
    \end{tabular}
\end{table}

\subsection{Optimization}
We first detail the optimization for monolingual tasks in Sec.~\ref{subsubsec:optim_deepsolo}. Then, we introduce the additionally introduced techniques for multilingual tasks in Sec.~\ref{subsubsec:optim_deepsolo++}.

\subsubsection{Optimization of DeepSolo}
\label{subsubsec:optim_deepsolo}
\noindent \textbf{Bipartite Matching.} After obtaining the prediction set $\hat{Y}$ and the ground truth (GT) set $Y$, similar to \citep{zhang2022text}, we use the Hungarian algorithm \citep{kuhn1955hungarian} to get an injective function $\varphi: [Y] \mapsto [\hat{Y}]$ that minimizes the matching cost $\mathcal{C}$:
\begin{equation}
    \underset{\varphi}{\arg \min} \sum\limits^{G-1}_{g=0}{\mathcal{C}(Y^{(g)}, \hat{Y}^{(\varphi(g))})}, \label{eq_4}
\end{equation}
where $G$ is the number of GT instances per image. 

Regarding the cost $\mathcal{C}$, previous work \citep{zhang2022text} only considers the class and position similarity while ignoring the similarity of the text script. However, matched pairs with similar positions may be quite different in texts, which could increase the optimization difficulty. TTS \citep{kittenplon2022towards} proposes a cross-entropy-based text matching criterion to address this issue, but it does not apply to our method since the character predictions are not aligned with the ground truth due to background class and repeated characters. We introduce a text matching criterion based on the typical Connectionist Temporal Classification (CTC) loss \citep{graves2006connectionist} which copes with the length inconsistency issue. For the $g$-th GT and its matched query, the complete cost function is:
\begin{equation}
\resizebox{.9\linewidth}{!}{$\begin{aligned}
    \mathcal{C}(Y^{(g)}, \hat{Y}^{(\varphi(g))}) = \lambda_{\text{cls}}\text{FL}'(\hat{b}^{(\varphi(g))}) + \lambda_{\text{text}} \text{CTC}(t^{(g)}, \hat{t}^{(\varphi(g))}) + \lambda_{\text{coord}} \sum_{n=0}^{N-1} \left\| p_n^{(g)} - \hat{p}_n^{(\varphi(g))}\right\|,
\end{aligned}$} \label{eq_5}
\end{equation}
where $\lambda_{\text{cls}}$, $\lambda_{\text{text}}$, and $\lambda_{\text{coord}}$ are hyper-parameters to balance different tasks. $\hat{b}^{(\varphi(g))}$ is the probability for the text-only instance class. The same as \citep{zhang2022text}, $\text{FL}'$ is defined as the difference between the positive and negative term: $\text{FL}'(x)=-\alpha (1-x)^\gamma \log(x) + (1-\alpha) x^\gamma \log(1-x)$, which is derived from the focal loss \citep{lin2017focal}. The second term is the CTC loss between the GT text $t^{(g)}$ and prediction $\hat{t}^{(\varphi(g))}$. The third term is the L1 distance between the GT $p_n^{(g)}$ and predicted point coordinates $\hat{p}_n^{(\varphi(g))}$ on the center curve.

\noindent \textbf{Overall Loss.} For the $k$-th query, the focal loss for instance classification is:
\begin{equation}
\resizebox{.9\linewidth}{!}{$\begin{aligned}
    \mathcal{L}_{\text{cls}}^{(k)} = -\mathds{1}_{\left\{ k \in \text{Im}(\varphi) \right\}} \alpha (1-\hat{b}^{(k)})^\gamma \log (\hat{b}^{(k)}) -\mathds{1}_{\left\{ k \notin \text{Im}(\varphi) \right\}} (1-\alpha) (\hat{b}^{(k)})^\gamma \log (1-\hat{b}^{(k)}),
\end{aligned}$}
\end{equation}
where $\mathds{1}$ is the indicator function, $\text{Im}(\varphi)$ is the image of the mapping $\varphi$. As for character classification, we exploit the CTC loss to address the length inconsistency issue between the GT text scripts and predictions:
\begin{equation}
    \mathcal{L}_{\text{text}}^{(k)} = \mathds{1}_{\left\{ k \in \text{Im}(\varphi) \right\}} \text{CTC}(t^{(\varphi^{-1}(k))}, \hat{t}^{(k)}).
\end{equation}
In addition, L1 distance loss is used for supervising point coordinates on the center curve and the boundaries (\ie, the top and bottom curves):
\begin{equation}
    \mathcal{L}_{\text{coord}}^{(k)} = \mathds{1}_{\left\{ k \in \text{Im}(\varphi) \right\}} \sum_{n=0}^{N-1} \left\|p_n^{(\varphi^{-1}(k))} - \hat{p}_n^{(k)} \right\|,
\end{equation}
\begin{equation}
\resizebox{.9\linewidth}{!}{$\begin{aligned}
\resizebox{1\linewidth}{!}{$
    \mathcal{L}_{\text{bd}}^{(k)} = \mathds{1}_{\left\{ k \in \text{Im}(\varphi) \right\}} \sum_{n=0}^{N-1} \left(\left\|top_n^{(\varphi^{-1}(k))} - \hat{top}_n^{(k)} \right\| + \left\|bot_n^{(\varphi^{-1}(k))} - \hat{bot}_n^{(k)} \right\|\right).
$}
\end{aligned}$}
\end{equation}

The loss function for the decoder consists of the four aforementioned losses:
\begin{equation}
    \mathcal{L}_{\text{dec}} = \sum_{k} \left( \lambda_{\text{cls}} \mathcal{L}_{\text{cls}}^{(k)}  + \lambda_{\text{text}} \mathcal{L}_{\text{text}}^{(k)} + \lambda_{\text{coord}} \mathcal{L}_{\text{coord}}^{(k)} + \lambda_{\text{bd}} \mathcal{L}_{\text{bd}}^{(k)}\right),
\end{equation}
where the hyper-parameters $\lambda_{\text{cls}}$, $\lambda_{\text{text}}$, and $\lambda_{\text{coord}}$ are the same as those in Eq.~(\ref{eq_5}). $\lambda_{\text{bd}}$ is the boundary loss weight. In addition, to make the Bezier center curve proposals introduced in Sec.~\ref{sec:top-k proposal} more accurate, we resort to adding intermediate supervision on the encoder. As we hope the points sampled on the Bezier center curve proposals are as close to the GT as possible, we calculate the L1 loss for the $N$ uniformly sampled points instead of only the four Bezier control points for each instance. This supervision method has been explored by \citep{feng2022rethinking}. The loss function for the encoder is formulated as:
\begin{equation}
    \mathcal{L}_{\text{enc}} = \sum_{i} \left( \lambda_{\text{cls}} \mathcal{L}_{\text{cls}}^{(i)}  + \lambda_{\text{coord}} \mathcal{L}_{\text{coord}}^{(i)}\right),
\end{equation}
where bipartite matching is also exploited to get one-to-one matching. The overall loss $\mathcal{L}$ is defined as:
\begin{equation}
    \mathcal{L} = \mathcal{L}_{\text{dec}} + \mathcal{L}_{\text{enc}}.
\end{equation}

\subsubsection{Optimization of DeepSolo++}
\label{subsubsec:optim_deepsolo++}

\begin{figure}[!t]
    \centering
    \includegraphics[width=0.8\linewidth]{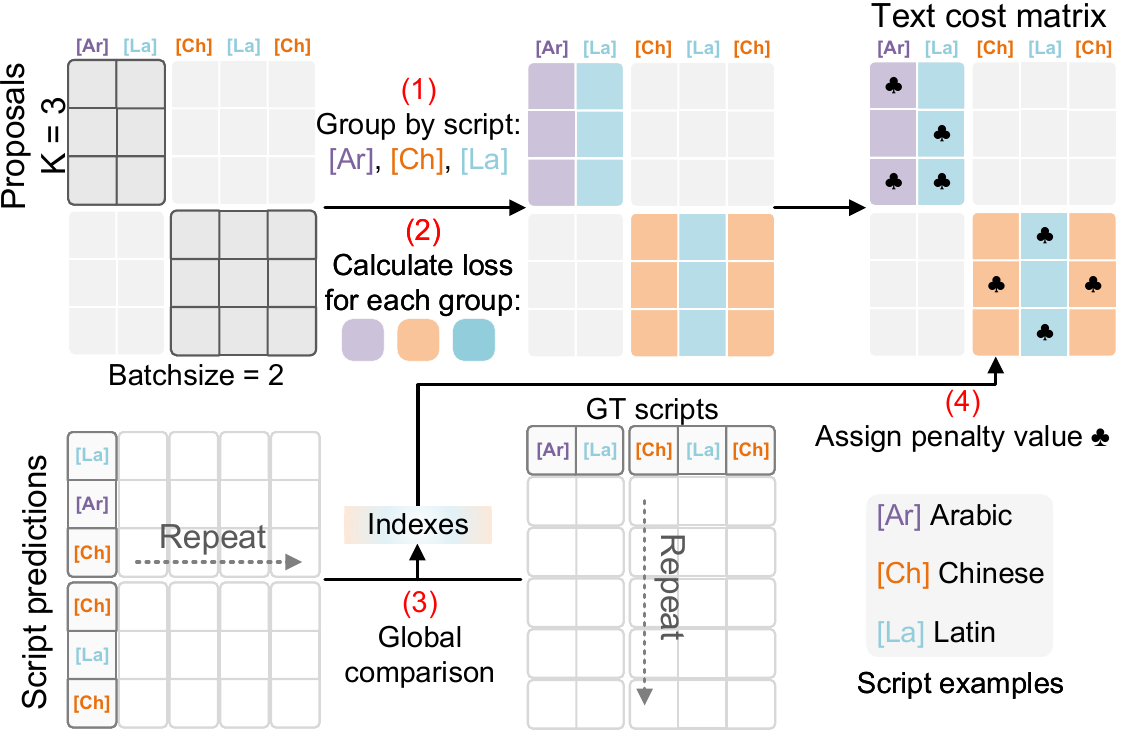}
    \caption{The illustration of script-aware matching.}
    \label{fig:lan_matching}
\end{figure}

\noindent\textbf{Script-aware Bipartite Matching.} 
In addition to the costs of the instance class, text transcript, and center point coordinates, the script identification cost (\ie, cross-entropy loss) is also involved in calculating the overall cost matrix. 
The hyper-parameter for balancing script identification cost is set to 1.0.
For text transcript cost, each batch may contain instances varying from different scripts, resulting in inconsistent character tables. Therefore, we cannot directly calculate the text loss using GT and the output of the selected linear layer according to script identification prediction.

To overcome this issue, we design the script-aware matching. The illustration is presented in Fig.~\ref{fig:lan_matching}. To be specific, our target in this period is to get a complete text cost matrix. Firstly, in each batch, we group the GT instances by script types. Next, for each script group, we select the output of the corresponding linear layer to calculate CTC text loss and fill the text cost matrix with the obtained text loss. Since gradients are not calculated during matching, $argmax$ can be used to obtain the script identification predictions. Then, we conduct a global comparison between the script identification prediction and GT. The indexes where the predicted script type is inconsistent with the GT can be obtained. We directly assign a penalty value, which is set to 20 by default, to these positions. Finally, we add text cost matrix and other cost matrices to achieve the overall one, which is used for bipartite matching.

\noindent\textbf{Overall Loss}. In addition to the losses of DeepSolo, the script identification loss is also included. We group the GT instances by script type when calculating the text loss. The final text loss is the summation of all script groups. Note that in some datasets, instances do not own their GT script types. It can be obtained by comparing transcripts with the character tables.

\section{Experiments}
\label{sec:exp}

\subsection{Monolingual Text Spotting}
\label{sec:experiments_deepsolo}
\subsubsection{Datasets and Evaluation Protocols}
We evaluate DeepSolo on \textbf{Total-Text} \citep{ch2020total}, \textbf{ICDAR 2015 (IC15)} \citep{karatzas2015icdar}, \textbf{SCUT-CTW1500} \citep{liu2019curved}, \textbf{ICDAR 2019 ReCTS} \citep{zhang2019icdar}, and \textbf{DAST1500} \citep{tang2019seglink++}. 
\textbf{Total-Text} is an arbitrarily shaped word-level benchmark, including 1,255 training images and 300 testing images. \textbf{IC15} contains 1,000 training images and 500 testing images. \textbf{SCUT-CTW1500} is a text-line level benchmark for long scene text with arbitrary shapes. There are 1,000 training images and 500 testing images. \textbf{ReCTS} is a benchmark for Chinese text on signboards, with 20,000 training images and 5,000 testing images. \textbf{DAST1500} is a benchmark for dense and long text, which contains 1,038 training images and 500 testing images.

For Total-Text, IC15, and SCUT-CTW1500, we adopt the following additional datasets for pre-training: 1) \textbf{Synth150K} \citep{liu2020abcnet}, 2) \textbf{ICDAR 2017 MLT} (MLT17) \citep{nayef2017icdar2017}, 3) \textbf{ICDAR 2013} (IC13) \citep{karatzas2013icdar}, 4) we also investigate the influence of \textbf{TextOCR}\citep{singh2021textocr}, which contains 21,778 training and 3,124 validation images.

For ReCTS, following previous works \citep{liu2021abcnet,huang2022swintextspotter,9960802}, the following datasets are used for pre-training: 1) \textbf{SynChinese130K} \citep{liu2021abcnet}, a synthetic dataset consists of about 130K images. 2) \textbf{ICDAR 2019 ArT} \citep{chng2019icdar2019}, which contains 5,603 training images. 3) \textbf{ICDAR 2019 LSVT} \citep{sun2019icdar}, a large-scale dataset where 30,000 images are used. 

We follow the classic protocols and adopt Precision (P), Recall (R), and H-mean (H) as metrics. H-mean is primary. For the end-to-end text spotting task on ReCTS, Normalized Edit Distance (NED) is adopted to calculate the 1-NED metric \citep{zhang2019icdar}.

\subsubsection{Implementation Details}  
The number of heads and sampling points for deformable attention is 8 and 4, respectively. The number of both encoder and decoder layers is 6. The number of proposals $K$ is 100. The number of sampled points $N$ is set to 25 for regular-length datasets, and otherwise 50 for the text-line level benchmark. Our models predict 38 character classes on Total-Text and IC15, 97 classes on CTW1500, and 5,463 classes on ReCTS (the background class is included). AdamW \citep{loshchilov2017decoupled} is used as the optimizer. The loss weights $\lambda_{\text{cls}}$, $\lambda_{\text{coord}}$,
$\lambda_{\text{bd}}$, and $\lambda_{\text{text}}$ are set to 1.0, 1.0, 0.5, and 0.5, respectively. For focal loss, $\alpha$ is 0.25 and $\gamma$ is 2.0.
The image batch size is 8. Following \citep{liao2021mask,zhong2021arts,huang2022swintextspotter}, data augmentations include random rotation, random crop, random scale, and color jittering.  Since there are no transcript labels in DAST1500, we abandon the character classification head and set $N$ to 8 for detection. More training details can be found in the appendix.

\begin{table*}[!t]
    \begin{minipage}[t]{0.5\linewidth}
        \caption{Effect of $\lambda_{\text{text}}$. `E2E' is the end-to-end spotting results, `None' refers to recognition without lexicon, and `Full' denotes recognition with a full lexicon that contains all in the test set.}
        \label{tab:ablation_loss}
        \centering
        \scalebox{0.6}{
            \begin{tabular}{c|ccc|cc}
            \toprule[1.1pt]
            \multirow{2}{*}{$\lambda_{\text{text}}$} & \multicolumn{3}{c|}{Detection} & \multicolumn{2}{c}{E2E} \\
            \cmidrule{2-4} \cmidrule{5-6} 
            & P & R & H & None & Full \\
            \midrule[1.1pt]
            0.25 & \textbf{94.29} & 82.07 & \textbf{87.76} & 76.68 & 85.76 \\
            \rowcolor{gray!20} \textbf{0.5} & 93.86 & \textbf{82.11} & 87.59 & \textbf{78.83} & \textbf{86.15} \\
            0.75 & 94.15 & 79.27 & 86.07 & 78.82 & 85.71 \\
            1.0 & 93.06 & 81.71 & 87.01 & 77.73 & 85.61 \\
            \bottomrule[1.1pt]
            \end{tabular}
        }
    \end{minipage}
    \hfill
    \begin{minipage}[t]{0.5\linewidth}
        \caption{Influence of sharing point embeddings and conducting text matching. The FPS is measured with 1 batch size on one A100 GPU.}
        \label{tab:ablation_sharing}
        \centering
        \vspace{3mm}
        \scalebox{0.6}{
            \setlength{\tabcolsep}{2pt}
            \begin{tabular}{cc|ccc|cc|c|c}
            \toprule[1.1pt]
            \multirow{2}{*}{Sharing} & \multirow{2}{*}{Matching} & \multicolumn{3}{c|}{Detection} & \multicolumn{2}{c|}{E2E} & \multirow{2}{*}{\#Params} & \multirow{2}{*}{FPS} \\ 
            \cmidrule{3-5} \cmidrule{6-7} 
            && P & R & H & None & Full & & \\ 
            \midrule[1.1pt]
            \rowcolor{gray!20} \ding{55} & \checkmark & \textbf{93.86} & \textbf{82.11} & \textbf{87.59} & \textbf{78.83} & \textbf{86.15} &42.5M &17.0 \\
            \checkmark & \checkmark & 93.60 & 81.21 & 86.96 & 77.58 & 85.98 &41.8M &17.0 \\
            \ding{55} & \ding{55} & 92.90 &82.11 &87.17 &77.85 &85.20 &42.5M &17.0\\
            \checkmark & \ding{55} &93.41 &81.98 &87.32 &77.09 &85.13 &41.8M &17.0 \\
            \bottomrule[1.1pt]
            \end{tabular}
        }
    \end{minipage}
\end{table*}

\subsubsection{Ablation Studies}
\label{sec:ablation}
We first conduct ablation studies on Total-Text, and then investigate the influence of training data and backbone.

\noindent \textbf{Text Loss Weight.} We study the influence of text loss weight, which has a direct impact on recognition performance. As shown in Tab.~\ref{tab:ablation_loss}, our model achieves a better trade-off between detection and end-to-end performance when $\lambda_{\text{text}}$ is set to 0.5. Thus we adopt $\lambda_{\text{text}} = 0.5$ for subsequent experiments.

\noindent \textbf{Sharing Point Embedding.} As shown in Tab.~\ref{tab:ablation_sharing}, when sharing point embeddings for all instances, the results on end-to-end task drop. It indicates that different instances require different point embeddings to encode the instance-specific features.

\noindent \textbf{Text Matching Criterion.} To evaluate the effectiveness of the text matching criterion, as shown in Tab.~\ref{tab:ablation_sharing}, we remove the text matching criterion from Eq.~(\ref{eq_5}). The primary end-to-end results decline. It validates the value of conducting text matching, which provides high-quality one-to-one matching according to both position and transcript similarity.

\noindent \textbf{Training Data.} We study the influence of different pre-training data in Tab.~\ref{tab:ablation_data}. For the end-to-end spotting task, with only Synth150K as the external pre-training data, our method can achieve 78.83\% accuracy without using a lexicon. With additional MLT17, IC13, and IC15 real scene data, the `None' and `Full' scores are improved by 0.82\% and 0.85\%, respectively. We further show that the performance is improved by a large margin using TextOCR. It demonstrates the value of using real data for pre-training and the scalability of our model on different data. In TextOCR, the average number of text instances per image is higher than in other datasets \citep{singh2021textocr}, which can provide more positive signals from the data perspective. Leveraging a real scene dataset with a larger scale and higher average number of instances may be more helpful for training DETR-like scene text detectors and spotters. On the other hand, in the DETR framework, one-to-one matching reduces the training efficiency of positive samples \citep{wang2022towards,jia2023detrs}. We speculate that the training efficiency of DeepSolo may also be affected by one-to-one matching. Improving the flaw of one-to-one matching while maintaining a low computational cost is worth further exploration.

\begin{table*}[!t]
    \begin{minipage}[t]{0.45\linewidth}
        \centering
        \caption{The influence of training data.}
        \label{tab:ablation_data}
        \vspace{2.5mm}
        \scalebox{0.5}{
        \setlength{\tabcolsep}{1pt}
        \begin{tabular}{c|l|c|ccc|cc}
        \toprule[1.1pt]
        \multirow{2}{*}{\#Row} &\multirow{2}{*}{External Data} &\multirow{2}{*}{Volume} & \multicolumn{3}{c|}{Detection} & \multicolumn{2}{c}{E2E} \\ 
        \cmidrule{4-6} \cmidrule{7-8} 
        & & & P & R & H & None & Full \\ 
        \midrule[1.1pt]
        1 & Synth150K & 150K & \textbf{93.86} & 82.11 & 87.59 & 78.83 & 86.15 \\
        2 & \#1+MLT17+IC13+IC15 & 160K & 93.09 & 82.11 & 87.26 & 79.65 & 87.00 \\
        \rowcolor{gray!20} 3 & \#2 +TextOCR & 185K & 93.19 & \textbf{84.64} & \textbf{88.72} & \textbf{82.54} & \textbf{88.72} \\
        \bottomrule[1.1pt]
        \end{tabular}
        }
    \end{minipage}
    \begin{minipage}[t]{0.55\linewidth}
        \centering
        \caption{Effect of different backbones. `Mem.': the peak memory of batching two images on one GPU.}
        \label{tab:ablation_backbone}
        \vspace{-1mm}
        \scalebox{0.6}{
        \setlength{\tabcolsep}{3pt}
        \begin{tabular}{l|ccc|cc|c|c}
        \toprule[1.1pt]
        \multirow{2}{*}{Backbone} & \multicolumn{3}{c|}{Detection} & \multicolumn{2}{c|}{E2E} & \multirow{2}{*}{\#Params} & \multirow{2}{*}{\makecell[c]{Mem.\\(MB)}} \\ 
        \cmidrule{2-4} \cmidrule{5-6} 
         & P & R & H & None & Full & & \\ 
        \midrule[1.1pt]
        ResNet-50 &93.09 &82.11 &87.26 &79.65 &87.00 &42.5M &17,216 \\
        Swin-T &92.77 &83.51 &87.90 &79.66 &87.05 &43.1M &26,573 \\
        ViTAEv2-S &92.57 &\textbf{85.50} &\textbf{88.89} &\textbf{81.79} &\textbf{88.40} &33.7M &25,332 \\
        \midrule
        ResNet-101 &93.20 &83.51 &88.09 &80.12 &87.14 &61.5M &19,541 \\
        Swin-S &\textbf{93.72} &84.24 &88.73 &81.27 &87.75 &64.4M &33,974 \\
        \bottomrule[1.1pt]
        \end{tabular}
        }
    \end{minipage}
\end{table*}

\begin{figure*}[!t]
    \begin{minipage}[t]{0.48\linewidth}
        \centering
        \includegraphics[width=1.0\linewidth]{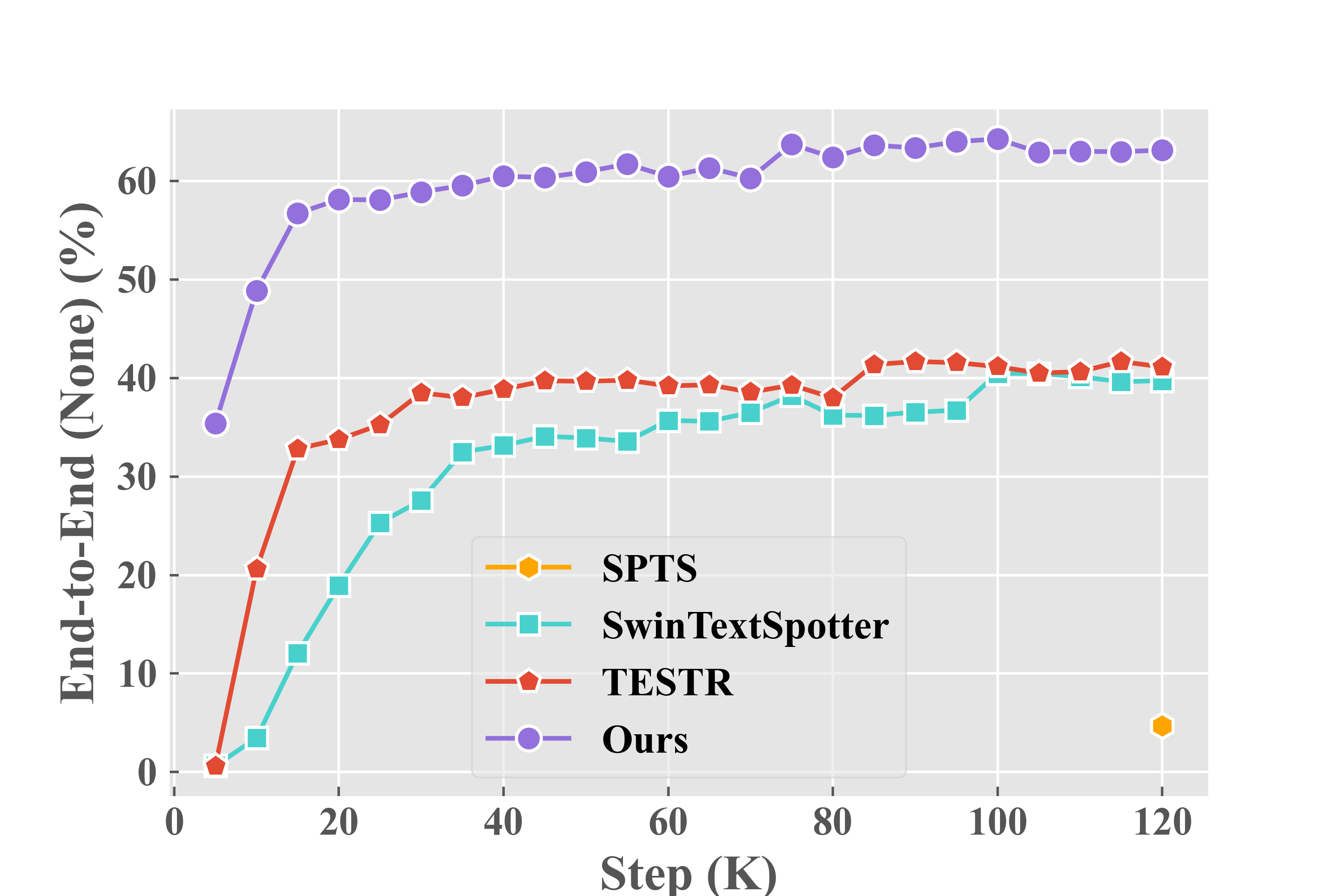}
        \caption{Comparison with open-sourced Transformer-based methods using only Total-Text training set.}
        \label{fig:train_eff_tt}
    \end{minipage}
    \hfill
    \begin{minipage}[t]{0.48\linewidth}
        \centering
        \includegraphics[width=1.0\linewidth]{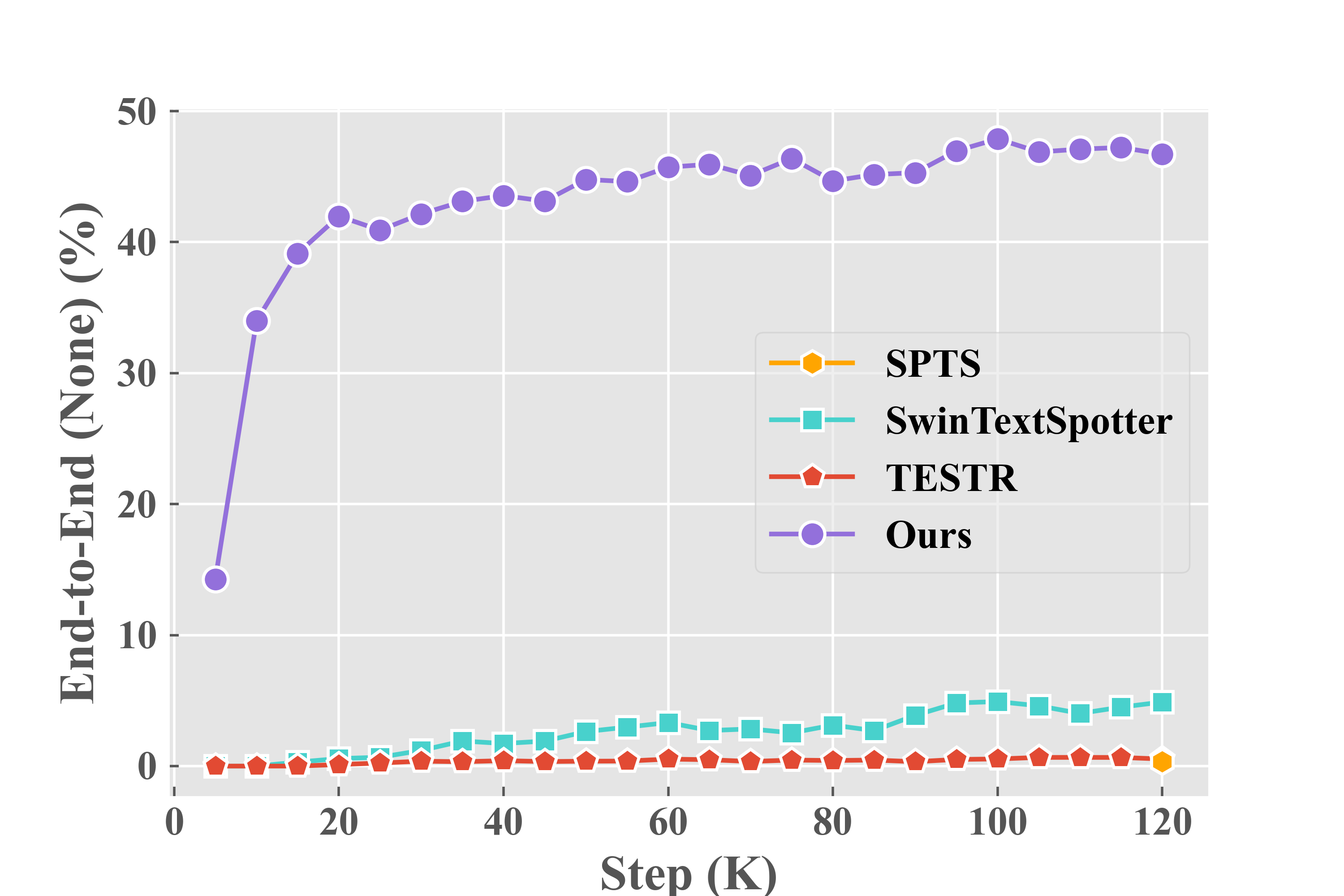}
        \caption{Comparison with open-sourced Transformer-based methods using only SCUT-CTW1500 training set.}
        \label{fig:train_eff_ctw}
    \end{minipage}
\end{figure*}

We also compare DeepSolo with existing open-source Transformer-based methods \citep{huang2022swintextspotter,zhang2022text,peng2022spts} by only using the training set of Total-Text. For a fair comparison, we apply the same data augmentation. Since both TESTR and our method are based on Deformable-DETR \citep{zhu2020deformable}, we set the same configuration for the Transformer modules. ResNet-50 \citep{he2016deep} is adopted in all experiments. The image batch size is set to 8, while the actual batch size of SPTS doubles due to batch augmentation. In Fig.~\ref{fig:train_eff_tt}, our method achieves faster convergence and better performance in the case of limited data volume, showing the superiority of DeepSolo over representative ones in training efficiency.

Taking a step further, we compare the training efficiency of our method with others on the long text benchmark, SCUT-CTW1500. We adopt the default recognition structural setting of all methods on SCUT-CTW1500. As shown in Fig.~\ref{fig:train_eff_ctw}, DeepSolo almost presents an emergent ability while using only 1,000 training images. A direct reason is that the `None' metric is nonlinear \citep{schaeffer2023emergent}. For long text detection and recognition, thanks to our proposed query form which contributes a unified representation, the two tasks are naturally aligned. Compared with TESTR which adopts dual decoders and heterogeneous queries, the results demonstrate the value of our proposed query form, which encodes a more accurate and explicit position prior and facilitates the learning of the single decoder. Consequently, our simpler design effectively mitigates the synergy issue and shows better training efficiency.

\begin{figure}[!t]
    \centering
    \includegraphics[width=\linewidth]{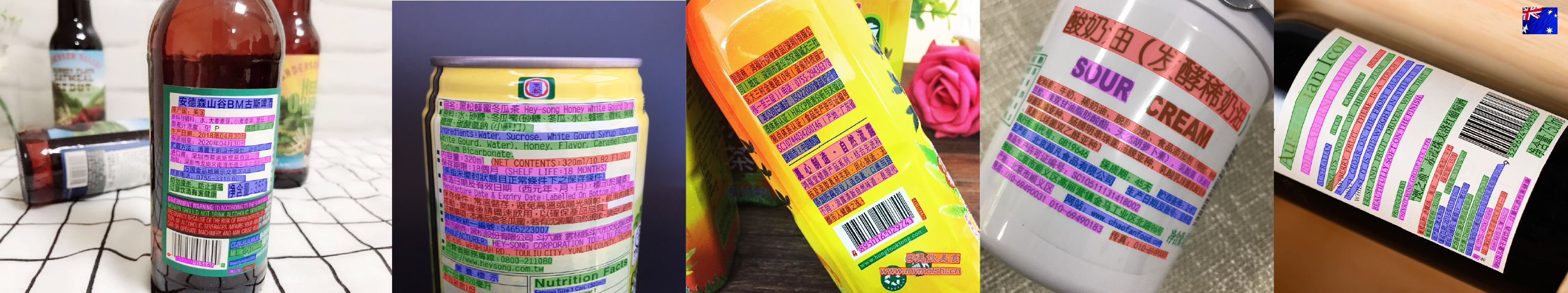}
    \vspace{-3mm}
    \caption{Qualitative detection results on DAST1500.}
    \label{fig:dast_vis}
\end{figure}

\begin{table*}[!t]
    \begin{minipage}[t]{0.5\linewidth}
        \centering
        \caption{Detection performance on DAST1500. `Ext.': extra external training data. `Res-50': ResNet-50. `\dag' denotes the results from \citep{tang2019seglink++}.}
        \label{tab:dast_detection}
        \scalebox{0.7}{
            \begin{tabular}{l|c|ccc}
            \toprule[1.1pt]
            Method &Ext. &P &R &H \\ 
            \midrule[1.1pt]
            SegLink \citep{shi2017detecting} \dag &\checkmark &66.0 &64.7 &65.3 \\
            CTD+TLOC \citep{liu2019curved} \dag &\checkmark &73.8 &60.8 &66.6 \\
            PixelLink \citep{deng2018pixellink} \dag &\checkmark &74.5 &75.0 &74.7 \\
            ICG \citep{tang2019seglink++} &\checkmark &79.6 &79.2 &79.4 \\
            ReLaText \citep{ma2021relatext} &\checkmark &\underline{89.0} &82.9 &85.8 \\
            MAYOR \citep{qin2021mask} &\ding{55} &87.8 &\underline{85.5} &\underline{86.6} \\
            \midrule
            \rowcolor{gray!20} DeepSolo (Res-50) &\ding{55} &\textbf{89.1} &\textbf{86.5}  &\textbf{87.8} \\
            \bottomrule[1.1pt]
            \end{tabular}
        }
    \end{minipage}
    \hfill
    \begin{minipage}[t]{0.5\linewidth}
        \centering
        \caption{Detection performance on SCUT-CTW1500. Methods in the top section do not adopt a DETR-like framework.}
        \label{tab:ctw_detection}
        \scalebox{0.6}{
            \begin{tabular}{l|c|ccc}
            \toprule[1.1pt]
            Method &Ext. &P &R &H \\ 
            \midrule[1.1pt]
            CRAFT \citep{baek2019character} &\checkmark &86.0 &81.1 &83.5 \\
            ContourNet \citep{wang2020contournet} &\ding{55} &83.7 &84.1 &83.9 \\
            PAN++ \citep{wang2021pan++} &\checkmark &87.1 &81.1 &84.0 \\
            DRRG \citep{zhang2020deep} &\checkmark &85.9 &83.0 &84.5 \\
            TextFuseNet \citep{ye2020textfusenet} &\checkmark &85.8 &85.0 &85.4 \\
            TextBPN++ (Res-50-DCN) \citep{zhang2022arbitrary} &\checkmark &88.3 &84.7 &86.5 \\
            FCENet (Res-50-DCN) \citep{zhu2021fourier} &\ding{55} &87.6 &83.4 &85.5 \\
            MAYOR \citep{qin2021mask} &\checkmark &88.7 &83.6 &86.1 \\
            DBNet++ \citep{liao2022real} &\checkmark &87.9 &82.8 &85.3 \\
            I3CL \citep{du2022i3cl} &\checkmark &88.4 &84.6 &86.5 \\
            TPSNet \citep{wang2022tpsnet} &\checkmark &88.7 &86.3 &87.5 \\
            \midrule
            Tang \etal \citep{tang2022few} &\checkmark &88.1 &82.4 &85.2 \\
            TESTR \citep{zhang2022text} &\checkmark &\underline{92.0} &82.6 &87.1 \\
            SwinTS (Swin-T) \citep{huang2022swintextspotter} &\checkmark &$-$ &$-$ &88.0 \\
            DPText-DETR \citep{ye2022dptext} &\checkmark &91.7 &\underline{86.2} &\underline{88.8} \\
            \rowcolor{gray!20} DeepSolo (Res-50) &\ding{55} &89.6 &85.8 &87.7 \\
            \rowcolor{gray!20} DeepSolo (Res-50) &\checkmark &\textbf{92.5} &\textbf{86.3} &\textbf{89.3} \\
            \bottomrule[1.1pt]
            \end{tabular}
        }
    \end{minipage}
\end{table*}

\noindent \textbf{Different Backbones.} We conduct experiments to investigate the influence of different backbones in Tab.~\ref{tab:ablation_backbone}. We select ResNet, Swin Transformer \citep{liu2021swin}, and ViTAEv2 \citep{zhang2023vitaev2} for comparison. All the models are pre-trained with the mixture data as listed in Row \#2 of Tab.~\ref{tab:ablation_data}. Compared with ResNet-50, ViTAEv2-S outperforms it by a large margin on the end-to-end spotting task, \ie, 2.14 \% on the `None' setting. Compared with ResNet-101, Swin-S achieves a gain of 1.15\% on `None' setting. We conjecture that the domain gap between large-scale synthetic data and small-scale real data might impact the performance of different backbones. It deserves further exploration to carefully tune the training configuration and study the training paradigm for text spotting.

\begin{table*}[!t]
\begin{minipage}[t]{1.0\linewidth}
    \centering
    \caption{Performance on Total-Text. `*': character-level annotations are used. `*' has the same meaning for other tables.}
    \label{tab:main_totaltext}
    \setlength{\tabcolsep}{4pt}
    \scalebox{0.58}{
    \begin{tabular}{l|l|ccc|cc|c|c}
    \toprule[1.1pt]
    \multirow{2}{*}{Method} & \multirow{2}{*}{External Data} & \multicolumn{3}{c|}{Detection} & \multirow{2}{*}{None} & \multirow{2}{*}{Full} & \multirow{2}{*}{\makecell[c]{FPS\\(report)}} & \multirow{2}{*}{\makecell[c]{FPS\\(A100)}} \\
    \cmidrule{3-5}
    & &P &R &H & & & & \\
    \midrule[1.1pt]
    
    TextDragon \citep{feng2019textdragon} &Synth800K &85.6 &75.7 &80.3 &48.8 &74.8 &$-$ &$-$ \\
    CharNet \citep{xing2019convolutional} $^{*}$ &Synth800K &88.6 &81.0 &84.6 &63.6 &$-$ &$-$ &$-$ \\
    TextPerceptron \citep{qiao2020text} &Synth800K &88.8 &81.8 &85.2 &69.7 &78.3 &$-$ &$-$\\
    CRAFTS \citep{baek2020character} $^{*}$ &Synth800K+IC13 &89.5 &85.4 &87.4 &78.7 &$-$ &$-$ &$-$ \\
    Boundary \citep{wang2020all} &Synth800K+IC13+IC15 &88.9 &85.0 &87.0 &65.0 &76.1 &$-$ &$-$ \\
    Mask TS v3 \citep{liao2020mask} &Synth800K+IC13+IC15+SCUT &$-$ &$-$ &$-$ &71.2 &78.4 &$-$ &$-$ \\
    PGNet \citep{wang2021pgnet} &Synth800K+IC15 &85.5 &\underline{86.8} &86.1 &63.1 &$-$ &35.5 &$-$ \\
    MANGO \citep{qiao2021mango} $^{*}$ &Synth800K+Synth150K+COCO-Text+MLT19+IC13+IC15 &$-$ &$-$ &$-$ &72.9 &83.6 &4.3 &$-$ \\
    PAN++ \citep{wang2021pan++} &Synth800K+COCO-Text+MLT17+IC15 &$-$ &$-$ &$-$ &68.6 &78.6 &21.1 &$-$ \\
    ABCNet v2 \citep{liu2021abcnet} &Synth150K+MLT17 &90.2 &84.1 &87.0 &70.4 &78.1 &10.0 &14.9 \\
    TPSNet (Res-50-DCN) \citep{wang2022tpsnet} &Synth150K+MLT17 &90.2 &\underline{86.8} &88.5 &76.1 &82.3 &9.3 &$-$ \\
    ABINet++ \citep{9960802} &Synth150K+MLT17 &$-$ &$-$ &$-$ &77.6 &84.5 &10.6 &$-$ \\
    GLASS \citep{ronen2022glass} &Synth800K &90.8 &85.5 &88.1 &79.9 &86.2 &3.0 &$-$ \\
    
    \midrule
    
    TESTR \citep{zhang2022text} &Synth150K+MLT17 &\underline{93.4} &81.4 &86.9 &73.3 &83.9 &5.3 &12.1 \\
    SwinTS (Swin-T) \citep{huang2022swintextspotter} &Synth150K+MLT17+IC13+IC15 &$-$ &$-$ &88.0 &74.3 &84.1 &$-$ &2.9 \\
    SPTS \citep{peng2022spts} &Synth150K+MLT17+IC13+IC15 &$-$ &$-$ &$-$ &74.2 &82.4 &$-$ &0.6 \\
    TTS (poly) \citep{kittenplon2022towards} &Synth800K+COCO-Text+IC13+IC15+SCUT &$-$ &$-$ &$-$ &78.2 &86.3 &$-$ &$-$ \\
    \rowcolor{gray!20} DeepSolo (Res-50) &Synth150K &\textbf{93.9} & 82.1 & 87.6 & 78.8 & 86.2 &17.0 &17.0 \\
    \rowcolor{gray!20} DeepSolo (Res-50) &Synth150K+MLT17+IC13+IC15 &93.1 & 82.1 & 87.3 & 79.7 & 87.0 &17.0 &17.0 \\
    \rowcolor{gray!20} DeepSolo (Res-50) &Synth150K+MLT17+IC13+IC15+TextOCR &93.2 & 84.6 & \underline{88.7} & \underline{82.5} & \underline{88.7} &17.0 &17.0 \\
    \rowcolor{gray!20} DeepSolo (ViTAEv2-S) &Synth150K+MLT17+IC13+IC15+TextOCR &92.9 & \textbf{87.4} & \textbf{90.0} & \textbf{83.6} & \textbf{89.6} &10.0 &10.0 \\
    \bottomrule[1.1pt]
    \end{tabular}
    }
\end{minipage}
\end{table*}

\begin{table*}[!t]
\begin{minipage}[t]{1.0\linewidth}
    \centering
    \caption{Performance on ICDAR2015. `S', `W', and `G' refer to using strong, weak, and generic lexicons, respectively.}
    \label{tab:main_ic15}
    \scalebox{0.52}{
    \begin{tabular}{l|l|ccc|ccc|ccc}
    \toprule[1.1pt]
    \multirow{2}{*}{Method} & \multirow{2}{*}{External Data} & \multicolumn{3}{c|}{Detection} & \multicolumn{3}{c|}{E2E} & \multicolumn{3}{c}{Word Spotting} \\
    \cmidrule{3-5} \cmidrule{6-8} \cmidrule{9-11} 
    & &P &R &H &S &W &G &S &W &G \\
    \midrule[1.1pt]
    
    TextDragon \citep{feng2019textdragon} &Synth800K &\underline{92.5} &83.8 &87.9 &82.5 &78.3 &65.2 &86.2 &81.6 &68.0 \\
    CharNet \citep{xing2019convolutional} $^{*}$ &Synth800K &91.2 &\underline{88.3} &89.7 &80.1 &74.5 &62.2 &$-$ &$-$ &$-$ \\
    TextPerceptron \citep{qiao2020text} &Synth800K &92.3 &82.5 &87.1 &80.5 &76.6 &65.1 &84.1 &79.4 &67.9 \\
    CRAFTS \citep{baek2020character} $^{*}$ &Synth800K+IC13 &89.0 &85.3 &87.1 &83.1 &82.1 &74.9 &$-$ &$-$ &$-$ \\
    Boundary \citep{wang2020all} &Synth800K+IC13+Total-Text &89.8 &87.5 &88.6 &79.7 &75.2 &64.1 &$-$ &$-$ &$-$ \\
    Mask TS v3 \citep{liao2020mask} &Synth800K+IC13+Total-Text+SCUT &$-$ &$-$ &$-$ &83.3 &78.1 &74.2 &83.1 &79.1 &75.1 \\
    PGNet \citep{wang2021pgnet} &Synth800K+Total-Text &91.8 &84.8 &88.2 &83.3 &78.3 &63.5 &$-$ &$-$ &$-$ \\
    MANGO \citep{qiao2021mango} $^{*}$ &Synth800K+Synth150K+COCO-Text+MLT19+IC13+Total-Text &$-$ &$-$ &$-$ &85.4 &80.1 &73.9 &85.2 &81.1 &74.6 \\
    PAN++ \citep{wang2021pan++} &Synth800K+COCO-Text+MLT17+Total-Text &$-$ &$-$ &$-$ &82.7 &78.2 &69.2 &$-$ &$-$ &$-$ \\
    ABCNet v2 \citep{liu2021abcnet} &Synth150K+MLT17 &90.4 &86.0 &88.1 &82.7 &78.5 &73.0 &$-$ &$-$ &$-$ \\
    ABINet++ \citep{9960802} &Synth150K+MLT17 &$-$ &$-$ &$-$ &84.1 &80.4 &75.4 &$-$ &$-$ &$-$ \\
    GLASS \citep{ronen2022glass} &Synth800K &86.9 &84.5 &85.7 &84.7 &80.1 &76.3 &86.8 &82.5 &78.8 \\
    \midrule
    TESTR \citep{zhang2022text} &Synth150K+MLT17+Total-Text &90.3 &\textbf{89.7} &\underline{90.0} &85.2 &79.4 &73.6 &$-$ &$-$ &$-$ \\
    SwinTS (Swin-T) \citep{huang2022swintextspotter} &Synth150K+MLT17+IC13+Total-Text &$-$ &$-$ &$-$ &83.9 &77.3 &70.5 &$-$ &$-$ &$-$ \\
    SPTS \citep{peng2022spts} &Synth150K+MLT17+IC13+Total-Text &$-$ &$-$ &$-$ &77.5 &70.2 &65.8 &$-$ &$-$ &$-$ \\
    TTS \citep{kittenplon2022towards} &Synth800K+COCO-Text+IC13+Total-Text+SCUT &$-$ &$-$ &$-$ &85.2 &81.7 &77.4 &85.0 &81.5 &77.3 \\

    \rowcolor{gray!20} DeepSolo (Res-50) &Synth150K+MLT17+IC13+Total-Text &\textbf{92.8} &87.4 &\underline{90.0} &86.8 &81.9 &76.9 &86.3 &82.3 &77.3 \\
    \rowcolor{gray!20} DeepSolo (Res-50) &Synth150K+MLT17+IC13+Total-Text+TextOCR &\underline{92.5} &87.2 &89.8 &\underline{88.0} &\underline{83.5} &\underline{79.1} &\underline{87.3} &\underline{83.8} &\underline{79.5} \\
    \rowcolor{gray!20} DeepSolo (ViTAEv2-S) &Synth150K+MLT17+IC13+Total-Text+TextOCR &92.4 &87.9 &\textbf{90.1} &\textbf{88.1} &\textbf{83.9} &\textbf{79.5} &\textbf{87.8} &\textbf{84.5} &\textbf{80.0} \\
    \bottomrule[1.1pt]
    \end{tabular}
    }
\end{minipage}
\end{table*}

\subsubsection{Comparison with State-of-the-art Methods}
\noindent \textbf{Dense and Long Text.}
DeepSolo adopts the Bezier center curve to represent scene text, rebuilds the text contour from center to boundary, and conducts explicit point refinement for final detection output. We show that our method effectively handles the detection of dense and long text. \textbf{1)} On DAST1500 (Tab.~\ref{tab:dast_detection}), even without pre-training, DeepSolo achieves the best performance, \ie, 87.8\% H-mean. Specifically, our method outperforms MAYOR by an absolute 1.2\% in terms of H-mean. Compared with ReLaText which adopts a synthetic dataset for pre-training, DeepSolo gets a 2.0\% higher H-mean score. Some qualitative detection results are shown in Fig.~\ref{fig:dast_vis}. \textbf{2)} On SCUT-CTW1500 (Tab.~\ref{tab:ctw_detection}), even without pre-training, DeepSolo achieves 87.7\% H-mean, outperforming all methods on the top section. Pre-trained on Synth150K, MLT17, Total-Text, IC13, and IC15, the H-mean is improved by 1.6\% to 89.3\%.

\begin{table*}[!t]
    \begin{minipage}[t]{0.49\linewidth}
        \centering
        \caption{End-to-end spotting result on SCUT-CTW1500. `len': the maximum recognition length. `\dag': measured on one A100 GPU.}
        \label{tab:main_ctw}
        \scalebox{0.6}{
        \begin{tabular}{l|cc|c}
        \toprule[1.1pt]
        Method & None & Full & FPS \\
        \midrule[1.1pt]
        TextDragon \citep{feng2019textdragon} &39.7 &72.4 &$-$ \\
        TextPerceptron \citep{qiao2020text} &57.0 &$-$ &$-$ \\
        ABCNet v2 (100 len) \citep{liu2021abcnet} &57.5 &77.2 &10.0 \\
        MANGO (25 len)\citep{qiao2021mango} &58.9 &78.7 &8.4 \\
        TPSNet(100 len, Res-50-DCN) \citep{wang2022tpsnet} &59.7 &79.2 &$-$ \\
        ABINet++ (100 len) \citep{9960802} &60.2 &80.3 &$-$ \\
        \midrule
        SwinTS (Swin-T) \citep{huang2022swintextspotter} &51.8 &77.0 &$-$ \\
        TESTR (100 len) \citep{zhang2022text} &56.0 &\underline{81.5} &15.9 \dag \\
        SPTS (100 len) \citep{peng2022spts} &\underline{63.6} &\textbf{83.8} &0.8 \dag \\
        \rowcolor{gray!20} DeepSolo (25 len, Res-50, Synth150K) &60.1 &78.4 &20.0 \dag \\
        \rowcolor{gray!20} DeepSolo (50 len, Res-50, Synth150K) &63.2 &80.0 &20.0 \dag \\
        \rowcolor{gray!20} DeepSolo (50 len, Res-50) &\textbf{64.2} &81.4 &20.0 \dag \\
        \bottomrule[1.1pt]
        \end{tabular}
        }
    \end{minipage}
    \hfill
    \begin{minipage}[t]{0.5\linewidth}
        \centering
        \caption{Detection and end-to-end recognition result (1-NED) on ReCTS.}
        \label{tab:main_rects}
        \vspace{3mm}
        \scalebox{0.7}{
        \begin{tabular}{l|ccc|c}
        \toprule[1.1pt]
        \multirow{2}{*}{Method} & \multicolumn{3}{c|}{Detection} & \multirow{2}{*}{1-NED} \\
        \cmidrule{2-4}
        &P &R &H & \\
        \midrule[1.1pt]
        FOTS \citep{liu2018fots} &78.3 &82.5 &80.3 &50.8 \\
        Mask TS v2 \citep{liao2021mask} &89.3 &88.8 &89.0 &67.8 \\
        AE TextSpotter \citep{wang2020ae} &92.6 &\textbf{91.0} &\textbf{91.8} &71.8 \\
        ABCNet v2 \citep{liu2021abcnet} &\underline{93.6} &87.5 &90.4 &62.7 \\
        ABINet++ \citep{9960802} &92.7 &89.2 &90.9 &76.5 \\
        \midrule
        SwinTS (Swin-T) \citep{huang2022swintextspotter} &\textbf{94.1} &87.1 &90.4 &72.5 \\
        \rowcolor{gray!20} DeepSolo (Res-50) &92.6 &89.0 &90.7 &\underline{78.3} \\
        \rowcolor{gray!20} DeepSolo (ViTAEv2-S) &92.6 &\underline{89.9} &\underline{91.2} &\textbf{79.6} \\
        \bottomrule[1.1pt]
        \end{tabular}
        }
    \end{minipage}
\end{table*}

\noindent \textbf{Results on Total-Text.} To evaluate the effectiveness of DeepSolo on scene text with arbitrary shape, we compare our model with state-of-the-art methods in Tab.~\ref{tab:main_totaltext}. \textbf{1)} Considering the `None' results, with only Synth150K as the external data, our method surpasses previous methods except GLASS. Compared with other Transformer-based methods, DeepSolo significantly outperforms TESTR, SwinTS, and SPTS by 5.5\%, 4.5\%, and 4.6\%, respectively. DeepSolo also outperforms TTS by 0.6\% while using far less training data. \textbf{2)} With additional MLT17, IC13, and IC15 real data, DeepSolo achieves 79.7\% in the `None' setting, which is comparable with the 79.9\% performance of GLASS. Note that DeepSolo runs much faster than GLASS and there is no elaborately tailored module for recognition, \eg, the Global to Local Attention Feature Fusion and the external recognizer in GLASS, while we only use a simple linear layer for recognition output. \textbf{3)} When using TextOCR, our method achieves very promising spotting performance, \ie, 82.5\% and 88.7\% at the `None' and `Full' settings. With ViTAEv2-S, the results are further improved by 1.1\% and 0.9\% in terms of `None' and `Full', respectively.

\noindent \textbf{Results on ICDAR 2015.} We conduct experiments on ICDAR 2015 to verify the effectiveness of DeepSolo on multi-oriented scene text, as presented in Tab.~\ref{tab:main_ic15}. The results show that DeepSolo achieves decent performance among all the comparing methods. Specifically, compared with Transformer-based methods using the same training datasets, DeepSolo surpasses SwinTS and SPTS by 6.4\% and 11.1\% on the E2E setting with the generic lexicon. With TextOCR, DeepSolo (ResNet-50) achieves the best `S', `W', and `G' metrics of 88.0\%, 83.5\%, and 79.1\%.

\noindent \textbf{Results on CTW1500.} In Tab.~\ref{tab:main_ctw}, pre-trained on Synth150K, DeepSolo with the maximum recognition length of 25 already outperforms most of the previous approaches on the `None' metric. We further increase the number of point queries from 25 to 50 for each text instance, achieving an absolute 3.1$\%$ improvement on the `None' result, without sacrificing inference speed. Following \citep{peng2022spts}, with MLT17, IC13, IC15, and Total-Text as the external training data, DeepSolo presents 64.2$\%$ H-mean without using lexicons, being 0.6$\%$ better and 25 times faster than SPTS.

\noindent \textbf{Results on ReCTS.} Chinese text owns thousands of character classes and far more complicated font structure than Latin, making Chinese text spotting a more difficult task. Following \citep{wang2020ae,liu2021abcnet,huang2022swintextspotter,9960802}, we evaluate the Chinese text spotting performance of DeepSolo on ReCTS. In Tab.~\ref{tab:main_rects}, we demonstrate that DeepSolo performs quite well compared with previous works, boosting the 1-NED metric to a new record of 78.3\%. It is worth noting that DeepSolo is not equipped with any explicit language modeling scheme but outperforms ABINet++ which leverages powerful language modeling by 1.8\% in terms of the 1-NED metric. Compared with SwinTS (Swin-T), DeepSolo (Res-50) achieves 5.8\% higher 1-NED performance. It demonstrates the inherent simplicity and remarkable extensibility of DeepSolo in tackling challenging language scenes.

\begin{table*}[!t]
    \begin{minipage}[t]{0.4\linewidth}
        \centering
        \caption{End-to-end spotting performance on RoIC13 without using lexicons.}
        \label{tab:roic13}
        \scalebox{0.55}{
        \begin{tabular}{l|ccc|ccc}
        \toprule[1.1pt]
        \multirow{2}{*}{Method} & \multicolumn{3}{c|}{Rotation 45$^\circ$} & \multicolumn{3}{c}{Rotation 60$^\circ$} \\
        \cmidrule{2-4} \cmidrule{5-7}
        &P &R &H &P &R &H \\
        \midrule[1.1pt]
        CharNet \citep{xing2019convolutional} &34.2 &35.5 &33.9 &10.3 &8.4 &9.3 \\
        Mask TS v2 \citep{liao2021mask} &66.4 &45.8 &54.2 &68.2 &48.3 &56.6 \\
        Mask TS v3 \citep{liao2020mask} &\textbf{88.5}  &66.8 &76.1 &\textbf{88.5} &67.6 &76.6 \\
        \midrule
        SwinTS (Swin-T) \citep{huang2022swintextspotter} &\underline{83.4} &\underline{72.5}  &\underline{77.6}  &\underline{84.6}  &\underline{72.1}  &\underline{77.9}  \\
        \rowcolor{gray!20} DeepSolo (Res-50) &82.3 &\textbf{74.9} &\textbf{78.4} &82.9 &\textbf{74.9} &\textbf{78.7} \\
        \bottomrule[1.1pt]
        \end{tabular}
        }
    \end{minipage}
    \begin{minipage}[t]{0.6\linewidth}
        \centering
        \caption{End-to-end spotting performance on Inverse-Text. `\#1', `\#2', and `\#3' denote the row index in Tab.~\ref{tab:ablation_data}.}
        \label{tab:inverse-text}
        \scalebox{0.6}{
        \begin{tabular}{l|cc}
        \toprule[1.1pt]
        Method &None &Full \\ 
        \midrule[1.1pt]
        Mask TS v2 \citep{liao2021mask} &39.0 &43.5 \\
        ABCNet \citep{liu2020abcnet} &22.2 &34.3 \\
        ABCNet v2 \citep{liu2021abcnet} &34.5 &47.4 \\
        \midrule
        TESTR \citep{zhang2022text} &34.2 &41.6 \\
        SwinTS (Swin-T) \citep{huang2022swintextspotter} &55.4 &67.9 \\
        SPTS \citep{peng2022spts} &38.3 &46.2 \\
        \rowcolor{gray!20} DeepSolo (Res-50, \#1) & 47.6 & 53.0 \\
        \rowcolor{gray!20} DeepSolo (Res-50, \#2) &48.5(\textcolor{blue}{+0.9}) &53.9(\textcolor{blue}{+0.9}) \\
        \rowcolor{gray!20} DeepSolo (Res-50, \#3) &\underline{64.6}(\textcolor{blue}{+17.0}) &\underline{71.2}(\textcolor{blue}{+18.2}) \\
        \rowcolor{gray!20} DeepSolo (ViTAEv2-S, \#3) &\textbf{68.8}(\textcolor{blue}{+21.2}) &\textbf{75.8}(\textcolor{blue}{+22.8}) \\
        \bottomrule[1.1pt]
        \end{tabular}
        }
    \end{minipage}
\end{table*}

\begin{figure}[!t]
    \centering
    \includegraphics[width=\linewidth]{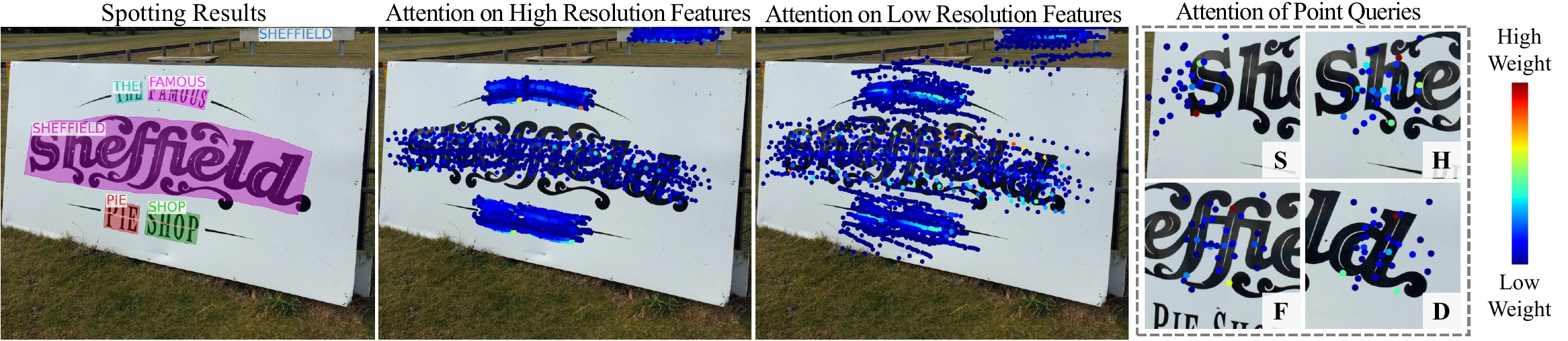}
    \vspace{-3mm}
    \caption{Visualization of the spotting results, attention on different scale features, and attention of point queries.}
    \label{fig:attn_vis}
\end{figure}

\begin{figure}[!t]
    \centering
    \includegraphics[width=\linewidth]{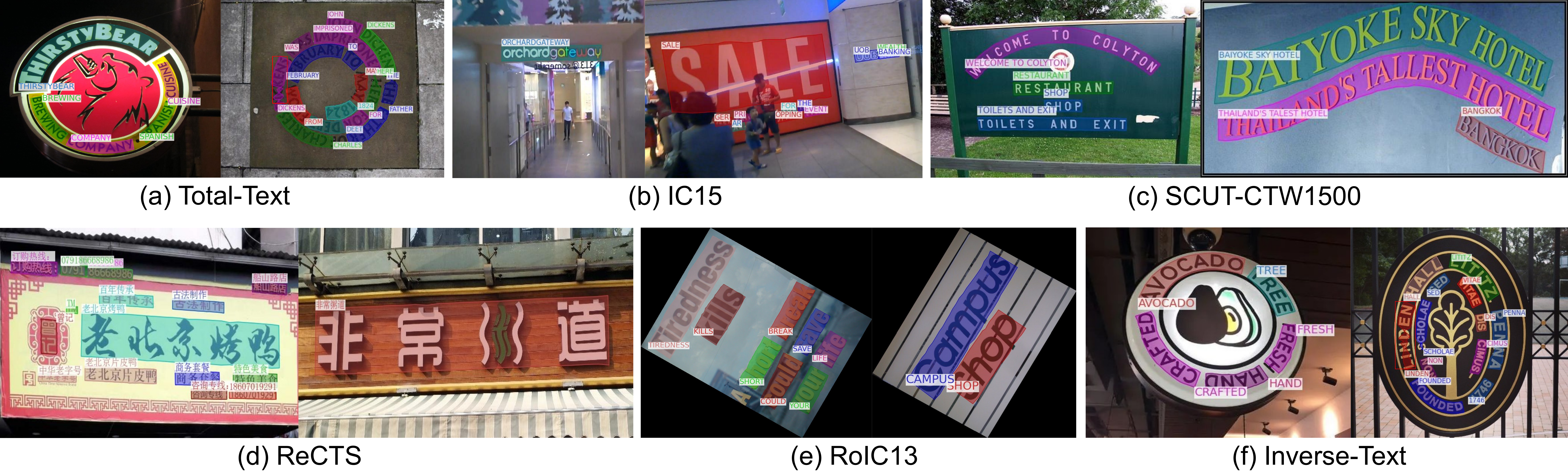}
    \vspace{-3mm}
    \caption{Qualitative results on six datasets. Two failure cases on vertically oriented texts are marked with \textcolor{red}{red} boxes.}
    \label{fig:vis_spotting}
\end{figure}

\subsubsection{Spotting Robustness on Rotated Text}
We conduct experiments on RoIC13 \citep{liao2020mask} and Inverse-Text \citep{ye2022dptext} to verify the rotation robustness of DeepSolo. On RoIC13, we fine-tune the pre-trained model (Tab.~\ref{tab:ablation_data} (\#2)) on IC13 using the rotation angles as \citep{liao2020mask,huang2022swintextspotter}. As shown in Tab.~\ref{tab:roic13}, DeepSolo achieves 78.4\% and 78.7\% H-mean in Rotation 45$^\circ$ and Rotation 60$^\circ$ settings.
Inverse-Text consists of 500 testing images in real scenes with about 40$\%$ inverse-like instances. The DeepSolo models reported in Tab.~\ref{tab:main_totaltext} are directly used for evaluation. The stronger rotation augmentation, \ie, using angle chosen from $[-90^{\circ}, +90^{\circ}]$ as in SwinTS, is not adopted. Results are shown in Tab.~\ref{tab:inverse-text}. When TextOCR is used, the `None' and `Full' performance are additionally improved by $16.1\%$ and $17.3\%$, respectively. While replacing ResNet-50 with ViTAEv2-S, $4.2\%$ and $4.6\%$ improvements on the two metrics are obtained.

\subsubsection{Visual Analysis}
Fig.~\ref{fig:attn_vis} visualizes the spotting results, the attention on different scale features, and the attention of different point queries. It shows that DeepSolo is capable of correctly recognizing scene texts of large size variance. 
In the rightest figure, it is noteworthy that point queries highly attend to the discriminative extremities of characters, which indicates that point queries can effectively encode the character position, scale, and semantic information. More qualitative results are presented in Fig.~\ref{fig:vis_spotting}. For the examples marked with red boxes, the recognition results are correct even though the detection results are invalid polygons.

\subsubsection{Compatibility to Line Annotations}
\begin{figure}[!t]
    \centering
    \includegraphics[width=0.8\linewidth]{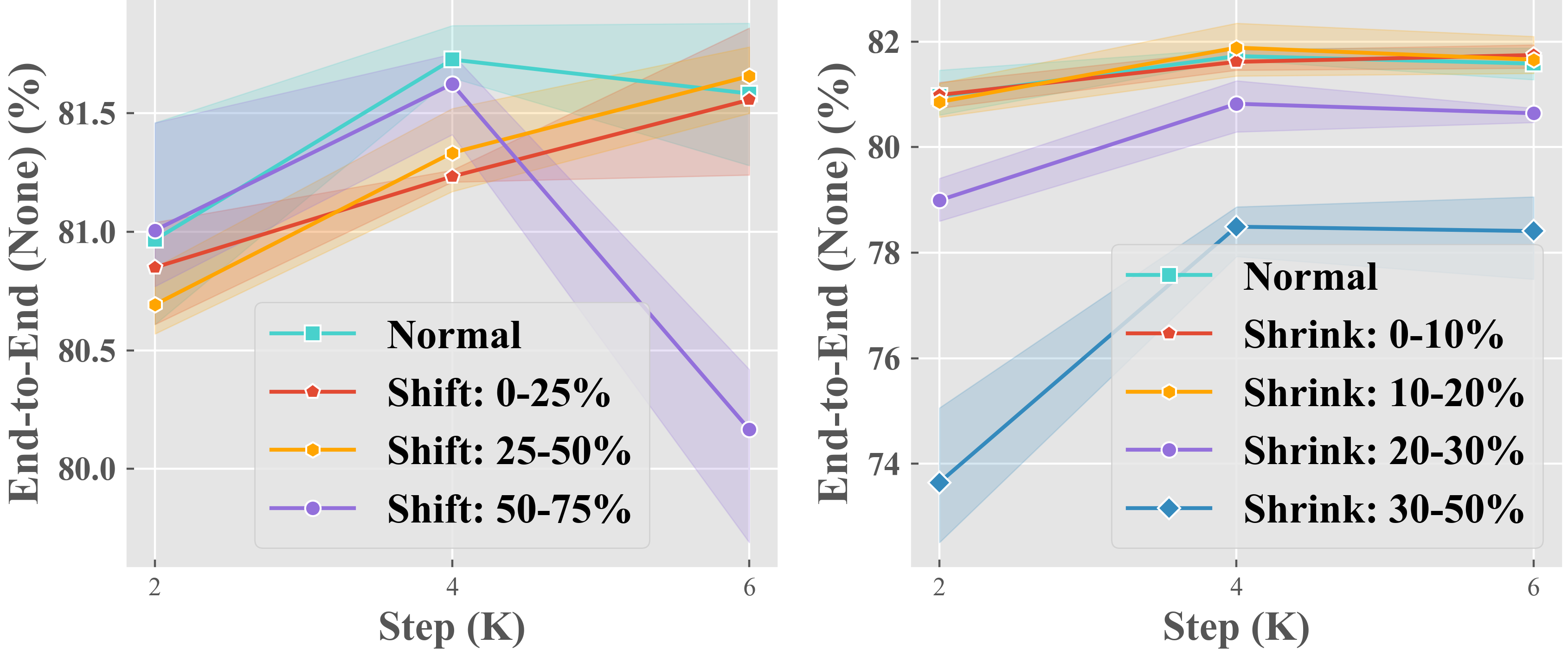}
    \caption{Analysis of the sensitivity to different line locations.}
    \label{fig:noise_analysis}
\end{figure}

DeepSolo can adapt to not only polygon annotations but also line annotations, which are much easier to obtain. We conduct experiments on Total-Text with only line annotations. To take advantage of existing full annotations, we first pre-train the model on a mixture of Synth150K, MLT17, IC13, IC15, and TextOCR. Then, we simply exclude the boundary head and fine-tune the model on Total-Text with IC13 and IC15 for 6K steps, using only the text center line annotations. During fine-tuning, the random crop augmentation which needs box information is discarded. We use the evaluation protocol provided by SPTS \citep{peng2022spts}. To further study the sensitivity to the line location, we randomly shift the center line annotations to the boundary and shrink them to the center point at different levels to simulate annotation errors. Results are plotted in Fig.~\ref{fig:noise_analysis}. It can achieve 81.6\% end-to-end (`None') performance, which is comparable with the fully supervised model, \ie, 82.5\%. As shown in Fig.~\ref{fig:noise_analysis}, the model is robust to the shift from 0\% to 50\% and the shrinkage from 0\% to 20\%. It indicates that the center line should not be too close to the text boundaries and better cover complete character areas. 

\subsection{Multilingual Text Spotting}
\label{sec:experiments_deepsolo_plus}
We validate the effectiveness of DeepSolo++ on ICDAR 2019 MLT \citep{nayef2019icdar2019} and ICDAR 2017 MLT \citep{nayef2017icdar2017}. We evaluate DeepSolo++ on three tasks, \ie, text localization (MLT19 Task 1, MLT17 Task 1, the task number follows the notification of official competition), joint text detection and script identification (MLT19 Task 3, MLT17 Task 3), and end-to-end text detection and recognition (MLT19 Task 4). In Sec.~\ref{sec:mlt_ablation}, we conduct ablation studies on MLT19 for DeepSolo++. Then, we compare DeepSolo++ with previous works on MLT19 and MLT17 in Sec.~\ref{sec:mlt_main_results}.

\subsubsection{Datasets and Evaluation Protocols}
\textbf{ICDAR 2017 MLT (MLT17)} \citep{nayef2017icdar2017} contains 7,200 training, 1,800 validation, and 9,000 testing images in 9 languages representing 6 different script types. Multi-oriented scene text instances are annotated with quadrilateral bounding boxes. \textbf{ICDAR 2019 MLT (MLT19)} \citep{nayef2019icdar2019} extends MLT17 for the end-to-end text recognition problem and adds a new language (Hindi) to the dataset. MLT19 contains 10,000 training and 10,000 testing images in 10 languages representing 7 different script types. A synthetic dataset (SynthTextMLT) \citep{buvsta2019e2e} containing 7 scripts is also provided for training.

Following \citep{Huang_2021_CVPR,huang2023task}, we also adopt the following datasets for pre-training: 1) SynthTextMLT, 2) MLT19, 3) ICDAR 2019 ArT \citep{chng2019icdar2019}, 4) ICDAR 2019 LSVT \citep{zhang2019icdar}, and 5) ICDAR 2017 RCTW \citep{shi2017icdar2017}, which contains 8,034 Chinese training images. Following the official competitions \citep{nayef2017icdar2017,nayef2019icdar2019}, evaluation protocols include Precision (P), Recall (R), H-mean (H), and Average Precision (AP).

\subsubsection{Implementation Details}
The number of sampled points $N$ is also set to 25. Other structural settings, such as the number of proposals, are the same as DeepSolo. For training the multilingual text spotter with multi-head routing structure, previous works \citep{Huang_2021_CVPR,huang2023task} initialize the weights using the released model of Mask TextSpotter v3 \citep{liao2021mask} and adopt three-stage training strategy, including training their recognizers individually. For DeepSolo++, we select the model weights in the first and third rows of Tab.~\ref{tab:ablation_data} for initialization. Then, we pre-train and fine-tune these models on MLT19. 

\begin{table*}[!h]
    \begin{minipage}[t]{0.55\linewidth}
        \centering
        \caption{Effect of position information for script token. `Task1': text detection. `Task3': joint text detection and script identification. `Task4': end-to-end text detection and recognition.}
        \label{tab:mlt19_ablation_token}
        \setlength{\tabcolsep}{2pt}
        \vspace{-1mm}
        \scalebox{0.8}{
            \begin{tabular}{c|cc|cc|ccc}
            \toprule[1.1pt]
            \multirow{2}{*}{Position} &\multicolumn{2}{c|}{Task1} & \multicolumn{2}{c|}{Task3} & \multicolumn{3}{c}{Task4} \\
            \cmidrule{2-3} \cmidrule{4-5} \cmidrule{6-8}
            &H &AP &H &AP &P &R &H\\ 
            \midrule
            Learnable &73.3 &61.5 &71.8 &59.9 &59.9 &39.2 &47.4 \\
            \midrule
            Start point &74.4 &63.4 &73.0 &61.7 &61.8 &41.3 &49.5 \\
            \rowcolor{gray!20} Center point &\textbf{74.8} &\textbf{63.9} &\textbf{73.4} &\textbf{62.3} &\textbf{62.5} &\textbf{42.0} &\textbf{50.3} \\
            \bottomrule[1.1pt]
            \end{tabular}
        }
    \end{minipage}
    \begin{minipage}[t]{0.45\linewidth}
        \centering
        \caption{The influence of text cost penalty in script-aware bipartite matching.}
        \label{tab:mlt19_ablation_penalty}
        \setlength{\tabcolsep}{2pt}
        \vspace{5mm}
        \scalebox{0.8}{
            \begin{tabular}{c|cc|cc|ccc}
            \toprule[1.1pt]
            \multirow{2}{*}{Penalty} &\multicolumn{2}{c|}{Task1} & \multicolumn{2}{c|}{Task3} & \multicolumn{3}{c}{Task4} \\
            \cmidrule{2-3} \cmidrule{4-5} \cmidrule{6-8}
            &H &AP &H &AP &P &R &H\\ 
            \midrule
            0 &75.6 &65.5 &74.3 &63.9 &61.0 &42.4 &50.0 \\
            5 &75.6 &65.4 &74.2 &63.7 &59.8 &42.0 &49.3 \\
            10 &75.8 &66.1 &74.3 &64.4 &59.5 &42.2 &49.4 \\
            20 &74.8 &63.9 &73.4 &62.3 &62.5 &42.0 &\textbf{50.3} \\
            \bottomrule[1.1pt]
            \end{tabular}
        }
    \end{minipage}
\end{table*}

\subsubsection{Ablation Studies}
\label{sec:mlt_ablation}
\noindent\textbf{The Positional Source for Script Token.} In Tab.~\ref{tab:mlt19_ablation_token}, we conduct experiments on comparing using explicit point information and learnable form \citep{carion2020end}. We find that using the explicit center point of each center curve achieves the better results. 

\noindent\textbf{The Text Cost Penalty in Script-aware Bipartite Matching.} 
As shown in Tab.~\ref{tab:mlt19_ablation_penalty}, we examine different penalty values in script-aware bipartite matching. Since we prefer the better end-to-end text detection and recognition (Task4) H-mean performance, we choose to assign a penalty value of 20 in the default setting.

\begin{table*}[!t]
    \begin{minipage}[t]{1.0\linewidth}
        \centering
        \caption{Text detection results on MLT19 with language-wise performance. The results of CRAFTS are from the official MLT19 website. CRAFTS (paper) denotes the results from their paper \citep{baek2020character}. The models in the top section are without a routing structure for multilingual recognition, and the following tables share the same format. `\#1' and `\#3' denote the row index in Tab.~\ref{tab:ablation_data}, where the pre-trained models are used for initialization.}
        \label{tab:mlt19_task1}
        \setlength{\tabcolsep}{3pt}
        \scalebox{0.7}{
            \begin{tabular}{l|cccc|ccccccc}
            \toprule[1.1pt]
            Method &P &R &H &AP &Arabic &Latin &Chinese &Japanese &Korean &Bangla &Hindi \\ 
            \midrule
            Single-head TextSpotter \citep{Huang_2021_CVPR} &83.8 &61.8 &71.1 &58.8 &51.1 &73.6 &40.4 &41.2 &56.5 &39.7 &49.0 \\
            CRAFTS \citep{baek2020character} &81.4 &62.7 &70.9 &56.6 &44.0 &72.5 &37.2 &42.1 &54.1 &38.5 &53.5 \\
            CRAFTS (paper) \citep{baek2020character} &81.7 &\textbf{70.1} &\underline{75.5} &$-$ &$-$ &$-$ &$-$ &$-$ &$-$ &$-$ &$-$ \\
            \midrule
            Multiplexed TextSpotter \citep{Huang_2021_CVPR} &85.5 &63.2 &72.7 &60.5 &51.8 &\underline{73.6} &43.9 &42.4 &57.2 &40.3 &52.0 \\
            \rowcolor{gray!20} DeepSolo++ (Res-50, routing, \#1) &\underline{85.8} &66.2 &74.8 &\underline{63.9} &\underline{53.1} &72.9 &\underline{49.7} &\underline{45.1} &\underline{62.0} &\underline{42.0} &\underline{55.7}\\
            \rowcolor{gray!20} DeepSolo++ (Res-50, routing, \#3) &\textbf{86.7} &\underline{68.2} &\textbf{76.3} &\textbf{66.1} &\textbf{54.8} &\textbf{74.9} &\textbf{50.5} &\textbf{45.5} &\textbf{62.7} &\textbf{42.5} &\textbf{56.0} \\
            \bottomrule[1.1pt]
            \end{tabular}
        }
    \end{minipage}
\end{table*}

\begin{table}[!t]
\centering
\caption{Text detection results on MLT17 compared with previous text spotting methods.}
\label{tab:mlt17_task1}
\begin{tabular}{l|ccc}
\toprule[1.1pt]
Method &P &R &H \\ 
\midrule
FOTS \citep{liu2018fots} &81.0 &57.5 &67.3 \\
CharNet \citep{xing2019convolutional} &77.1 &\textbf{70.1} &73.4 \\
\midrule
Multiplexed TextSpotter \citep{Huang_2021_CVPR} &\underline{85.4} &62.9 &72.4 \\
\rowcolor{gray!20} DeepSolo++ (Res-50, routing, \#1) &85.3 &65.2 &\underline{73.9} \\
\rowcolor{gray!20} DeepSolo++ (Res-50, routing, \#3) &\textbf{86.2} &\underline{67.2} &\textbf{75.6} \\
\bottomrule[1.1pt]
\end{tabular}
\end{table}

\subsubsection{Comparison with State-of-the-art Methods}
\label{sec:mlt_main_results}
\noindent\textbf{Text Detection Task.} As listed in Tab.~\ref{tab:mlt19_task1} and Tab.~\ref{tab:mlt17_task1}, DeepSolo++ surpasses \citep{Huang_2021_CVPR} by a clear margin. It is noteworthy that DeepSolo++ significantly outperforms previous methods in Chinese, Japanese, and Korean text.

\begin{table*}[!h]
    \begin{minipage}[t]{0.49\linewidth}
        \centering
        \caption{Joint text detection and script identification results on MLT19.}
        \label{tab:mlt19_task3}
        \setlength{\tabcolsep}{2pt}
        \scalebox{0.7}{
        \begin{tabular}{l|cccc}
        \toprule[1.1pt]
        Method &P &R &H &AP \\ 
        \midrule
        Single-head TextSpotter \citep{Huang_2021_CVPR} &75.4 &57.4 &65.2 &52.0 \\
        CRAFTS \citep{baek2020character} &78.5 &60.5 &68.3 &53.8 \\
        \midrule
        Multiplexed TextSpotter \citep{Huang_2021_CVPR} &81.7 &60.3 &69.4 &56.5 \\
        \rowcolor{gray!20} DeepSolo++ (Res-50, routing, \#1) &\underline{84.3} &\underline{65.0} &\underline{73.4} &\underline{62.3} \\
        \rowcolor{gray!20} DeepSolo++ (Res-50, routing, \#3) &\textbf{85.1} &\textbf{66.9} &\textbf{74.9} &\textbf{64.5} \\
        \bottomrule[1.1pt]
        \end{tabular}
        }
    \end{minipage}
    \hfill
    \begin{minipage}[t]{0.49\linewidth}
        \centering
        \caption{Joint text detection and script identification results on MLT17.}
        \label{tab:mlt17_task3}
        \setlength{\tabcolsep}{2pt}
        \scalebox{0.7}{
        \begin{tabular}{l|cccc}
        \toprule[1.1pt]
        Method &P &R &H &AP \\ 
        \midrule
        E2E-MLT \citep{buvsta2019e2e} &64.6 &53.8 &58.7 &$-$ \\
        CRAFTS \citep{baek2020character} &74.5 &63.1 &68.3 &54.6 \\
        \midrule
        Multiplexed TextSpotter \citep{Huang_2021_CVPR} &81.8 &60.3 &69.4 &56.3 \\
        \rowcolor{gray!20} DeepSolo++ (Res-50, routing, \#1) &\underline{83.7} &\underline{64.0} &\underline{72.5} &\underline{61.1} \\
        \rowcolor{gray!20} DeepSolo++ (Res-50, routing, \#3) &\textbf{84.6} &\textbf{65.9} &\textbf{74.1} &\textbf{63.3} \\
        \bottomrule[1.1pt]
        \end{tabular}
        }
    \end{minipage}
\end{table*}

\noindent\textbf{Joint Text Detection and Script Identification Task.} In Tab.~\ref{tab:mlt19_task3} and Tab.~\ref{tab:mlt17_task3}, without a specific language prediction network, DeepSolo++ shows extraordinary end-to-end script identification ability with a single script token. On MLT19, compared with \citep{Huang_2021_CVPR}, DeepSolo++ (Res-50, routing, \#1) achieves 4.0\% higher H-mean and 5.8\% higher AP score. DeepSolo++ (Res-50, routing, \#3) achieves 5.5\% and 8.0\% improvement in terms of H-mean and AP, respectively.

\noindent\textbf{End-to-End Recognition Task.} As shown in Tab.~\ref{tab:mlt19_task4}, DeepSolo++ (Res-50, routing, \#3) achieves 51.2\% H-mean performance, being 2.7\% higher than Grouped TextSpotter. However, we notice that there is still a gap between the routing models in the bottom section and the method using a single recognizer, \ie, CRAFTS (paper). On the one hand, CRAFTS only predicts 4,267 character classes and additionally uses 20,000 images of ReCTS for training \citep{baek2020character}. Besides, they use a heavy recognizer which is with 24 Conv layers. On the other hand, the text loss is sparse for all linear layers in our model which are responsible for character classification. Since each batch may not contain all script types, all linear layers cannot be trained together in every step, which might affect the training efficiency.

\begin{figure}[!t]
    \centering
    \includegraphics[width=\linewidth]{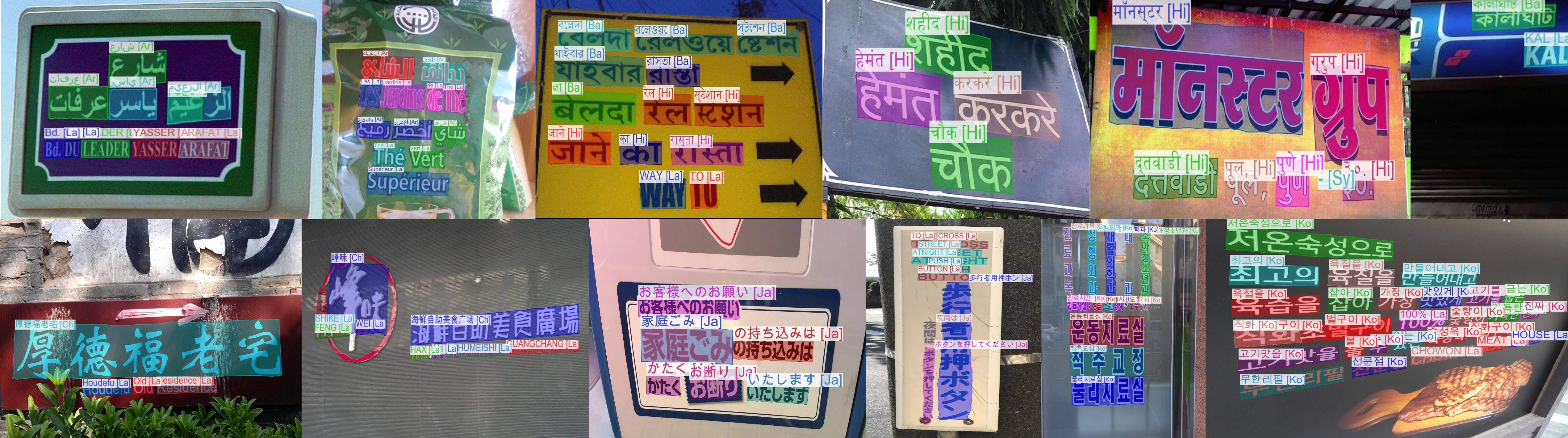}
    \caption{Text spotting visualization results on ICDAR 2019 MLT.}
    \label{fig:vis_mlt}
\end{figure}

\begin{table}[!t]
    \centering
    \caption{End-to-end recognition results on MLT19.}
    \label{tab:mlt19_task4}
    \begin{tabular}{l|ccc}
    \toprule[1.1pt]
    Method &P &R &H \\ 
    \midrule
    E2E-MLT \citep{buvsta2019e2e}  &37.4 &20.5 &26.5 \\
    RRPN+CLTDR \citep{ma2018arbitrary} &38.6 &30.1 &33.8 \\
    Single-head TextSpotter \citep{Huang_2021_CVPR} &\underline{71.8} &27.4 &39.7 \\
    CRAFTS \citep{baek2020character} &65.7 &42.7 &\underline{51.7} \\
    CRAFTS (paper) \citep{baek2020character} &\textbf{72.9} &\textbf{48.5} &\textbf{58.2} \\
    \midrule
    Multiplexed TextSpotter \citep{Huang_2021_CVPR} &68.0 &37.3 &48.2 \\
    Grouped TextSpotter (5 heads) \citep{huang2023task} &67.7 &37.8 &48.5 \\
    \rowcolor{gray!20} DeepSolo++ (Res-50, routing, \#1) &62.5 &42.0 &50.3 \\
    \rowcolor{gray!20} DeepSolo++ (Res-50, routing, \#3) &62.3 &\underline{43.5} &51.2 \\
    \bottomrule[1.1pt]
    \end{tabular}
\end{table}

\section{Limitation and Discussion}
\label{sec:discussion}
Although our method demonstrates promising performance on extensive datasets, there is still some room for improvement. \textbf{1)} The point order in annotations is in line with the reading order, which implicitly guides the model to learn the text order. When the position annotation is not in line with the reading order or the predicted order is incorrect, how to get robust detection \citep{ye2022dptext} and correct recognition is worth further exploration. 
\textbf{2)} In DeepSolo++, we model the queries for different languages in a single Transformer decoder. Whether it is necessary to determine a solution to obtain discriminative features for different languages is worth further exploration.
\textbf{3)} We do not leverage explicit language modeling. The combination of DETR-based DeepSolo and language modeling may be promising. 
\textbf{4)} The long-tail recognition issue is still not fixed yet in cutting-edge spotters.

\section{Conclusion}
\label{sec:conclusion}
In this paper, we propose simple yet effective baseline models for monolingual and multilingual text spotting, taking the merit of a novel explicit point query form that provides a pivotal representation for different tasks. With a single decoder and several simple prediction heads, we present a much simpler method compared with previous text spotters. Our method shows several good properties, including 1) simplicity of structure and training pipeline, 2) efficiency of training and inference, and 3) extensibility of character class, language, and task. Extensive experiments demonstrate that our method has achieved SOTA performance while enjoying some other distinctions, such as the effectiveness on dense and long text, and compatibility to line annotations.

\section*{Acknowledgements}

This work was supported in part by the National Natural Science Foundation of China under Grants U23B2048, 62076186 and 62225113, and in part by the National Key Research and Development Program of China under Grant 2023YFC2705700. The numerical calculations in this paper have been done on the supercomputing system in the Supercomputing Center of Wuhan University.

\begin{appendices}

\section{Details of Training}
The training details of DeepSolo with ResNet \citep{he2016deep} (ImageNet pre-trained weights from TORCHVISION) are listed in Tab.~\ref{tab:resnet} with corresponding training data, learning rate, and iterations. 
In Fig. 5 and Fig. 6 of the main paper, \ie, only the Total-Text or CTW1500 training set is utilized, the training schedule of DeepSolo is related to Row \#9 and Row \#10, respectively. For SPTS \citep{peng2022spts}, we only plot the final performance since SPTS needs more data and a longer training schedule to achieve ideal performance. 
The training setting of DeepSolo with line labels is provided in Row \#12. During fine-tuning, the line annotations are used and stronger rotation augmentation (angle randomly chosen from $[-90^{\circ}, +90^{\circ}]$) is adopted.

\begin{table*}[h]
\begin{minipage}[t]{1.0\linewidth}
    \centering
    \caption{Training details of DeepSolo with ResNet. `Step' denotes the iteration step where the learning rate is divided by 10.}
    \label{tab:resnet}
    \setlength{\tabcolsep}{2pt}
    \scalebox{0.42}{
    \begin{tabular}{c|l|lccc|lccc|l}
    \toprule[1.1pt]
    \multirow{2}{*}{\#Row} & \multirow{2}{*}{Backbone} & \multicolumn{4}{c|}{Pre-training} & \multicolumn{4}{c|}{Fine-tuning} & \multirow{2}{*}{Where in the Main Paper} \\
    \cmidrule{3-6} \cmidrule{7-10}
    & &Training Data &$lr$ (Backbone) &Iterations &Step &Training Data &$lr$ (Backbone) &Iterations &Step & \\
    \midrule[1.1pt]
    1 &\multirow{7}{*}{ResNet-50} &Synth150K+Total-Text &$1e^{-4}$ ($1e^{-5}$) &350K &300K &Total-Text &$1e^{-5}$ ($1e^{-6}$) &10K &$-$ &Tab. 2, 3, 4, 8\\
    2 & &Synth150K+Total-Text+MLT17+IC13+IC15 &$1e^{-4}$ ($1e^{-5}$) &375K &320K &Total-Text &$1e^{-5}$ ($1e^{-6}$) &10K &$-$ &Tab. 4, 5, 8\\
    3 & &Synth150K+Total-Text+MLT17+IC13+IC15+TextOCR &$1e^{-4}$ ($1e^{-5}$) &435K &375K &Total-Text &$1e^{-5}$ ($1e^{-6}$) &2K &$-$ &Tab. 4, 8\\
    \cmidrule{3-11}
    4 & &Synth150K+Total-Text+MLT17+IC13+IC15 &$1e^{-4}$ ($1e^{-5}$) &375K &320K &IC15 &$1e^{-5}$ ($1e^{-6}$) &3K &$-$ &Tab. 9\\
    5 & &Synth150K+Total-Text+MLT17+IC13+IC15+TextOCR &$1e^{-4}$ ($1e^{-5}$) &435K &375K &IC15 &$1e^{-5}$ ($1e^{-6}$) &1K &$-$ &Tab. 9 \\
    \cmidrule{3-11}
    6 & &Synth150K+Total-Text+MLT17+IC13+IC15 &$1e^{-4}$ ($1e^{-5}$) &375K &320K &CTW1500 &$5e^{-5}$ ($5e^{-6}$) &12K &8K &Tab. 7, 10\\
    \cmidrule{3-11}
    7 & &SynChinese130K+ArT+LSVT+ReCTS &$1e^{-4}$ ($1e^{-5}$) &400K &300K &ReCTS &$1e^{-5}$ ($1e^{-6}$) &30K &20K &Tab. 11\\
    \cmidrule{3-11}
    8 & &Synth150K+Total-Text+MLT17+IC13+IC15 &$1e^{-4}$ ($1e^{-5}$) &375K &320K &IC13 &$1e^{-5}$ ($1e^{-6}$) &1K &$-$ &Tab. 12\\
    \cmidrule{3-11}
    9 & &Total-Text &$1e^{-4}$ ($1e^{-5}$) &120K &80K &$-$ &$-$ &$-$ &$-$ &Fig. 5\\
    10 & &CTW1500 &$1e^{-4}$ ($1e^{-5}$) &120K &90K &$-$ &$-$ &$-$ &$-$ &Fig. 6, Tab. 7\\
    11 & &DAST1500 &$1e^{-4}$ ($1e^{-5}$) &80K &70K, 78K &$-$ &$-$ &$-$ &$-$ &Tab. 6\\
    \cmidrule{3-11}
    12 & &Synth150K+MLT17+IC13+IC15+TextOCR &$1e^{-4}$ ($1e^{-5}$) &435K &375K &Total-Text+IC13+IC15 &$2e^{-5}$ ($2e^{-6}$) &6K &$-$ &Fig. 10\\
    \cmidrule{3-11}
    13 & &SynthTextMLT+MLT19+ArT+LSVT+RCTW &$1e^{-4}$ ($1e^{-5}$)  &600K &450K &MLT19 &$1e^{-5}$ ($1e^{-6}$) &50K &$-$ &Tab. 14-20 \\
    \midrule
    14 &\multirow{1}{*}{ResNet-101} &Synth150K+Total-Text+MLT17+IC13+IC15 &$1e^{-4}$ ($1e^{-5}$) &375K &320K &Total-Text &$1e^{-5}$ ($1e^{-6}$) &10K &$-$ &Tab. 5\\
    \bottomrule[1.1pt]
    \end{tabular}
    }
\end{minipage}
\end{table*}

In Tab. 5 of the main paper, with Swin Transformer \citep{liu2021swin}, we pre-train the model for 375K iterations and fine-tune it on Total-Text for 10K iterations. No part of the backbone is frozen. During pre-training, the initial learning rate for the backbone is $1e^{-4}$. The drop path rate of Swin-T and Swin-S is set to 0.2 and 0.3, respectively. During fine-tuning, we set the learning rate for the backbone to $1e^{-5}$, and the drop path rate to 0.2 and 0.3 for Swin-T and Swin-S. Other training schedules are the same as Row \#2 in Tab.~\ref{tab:resnet}.

With ViTAEv2-S \citep{zhang2023vitaev2}, the drop path rate is set to 0.3 for pre-training and 0.2 for fine-tuning. Other training schedules are the same as Swin Transformer backbone.

\begin{figure}[!t]
    \centering
    \includegraphics[width=0.9\linewidth]{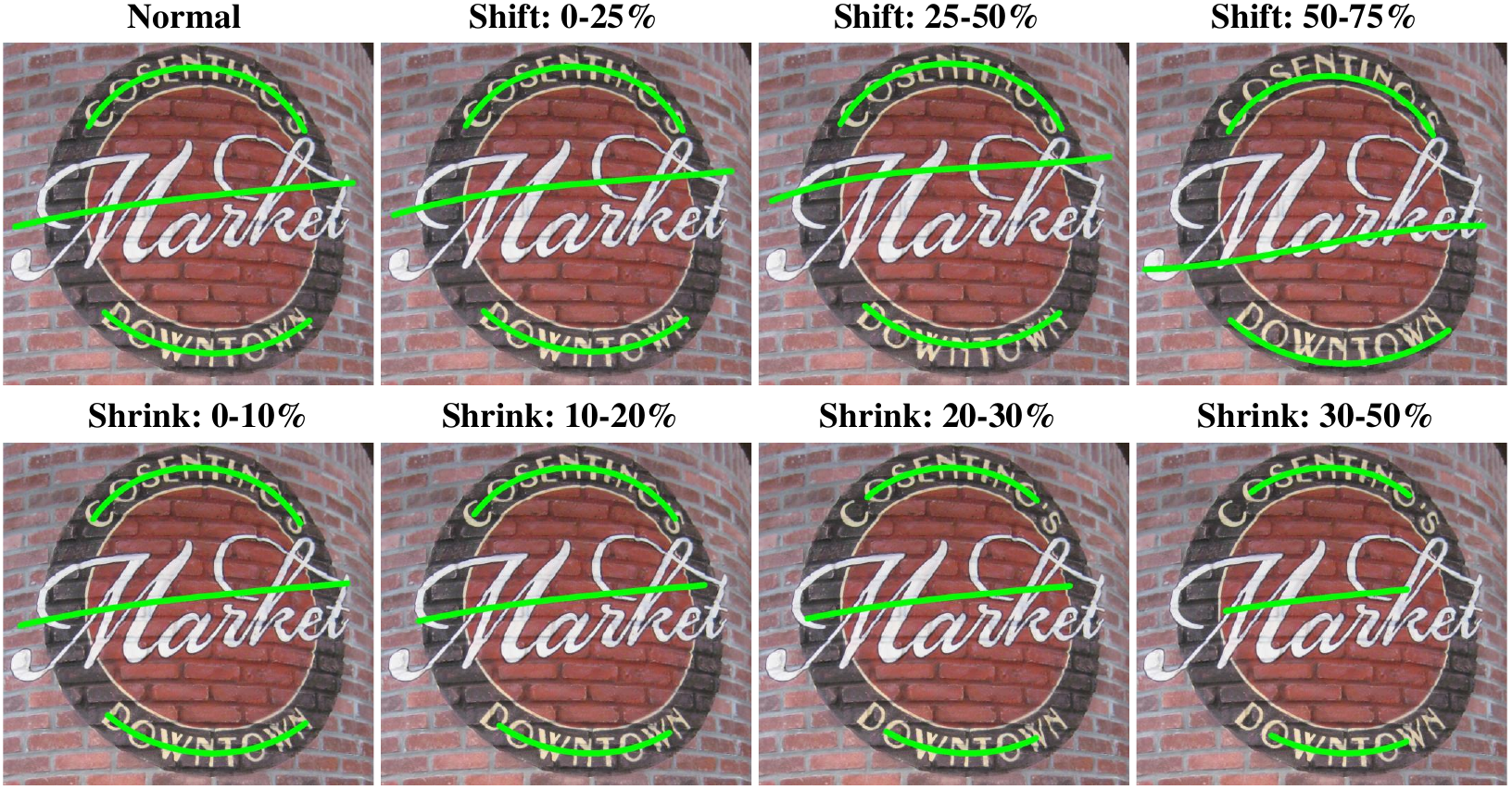}
    \caption{The illustration of line labels at different noisy levels.}
    \label{fig:noise_label_vis}
\end{figure}

\begin{figure}[!t]
    \centering
    \includegraphics[width=0.9\linewidth]{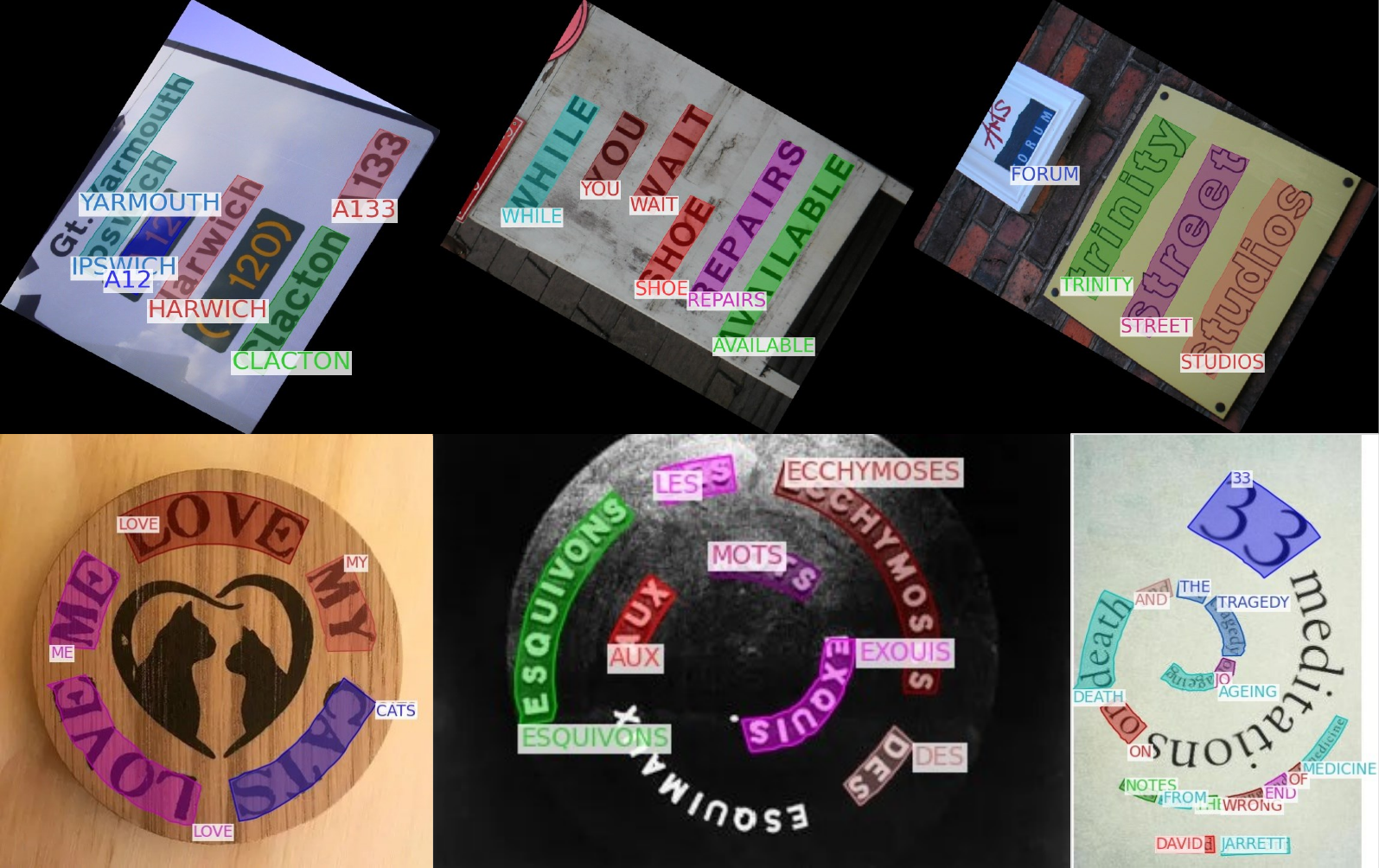}
    \caption{Spotting visualizations on RoIC13 (the first row, ResNet-50 backbone) and Inverse-Text (the bottom row, ViTAEv2-S backbone).}
    \label{fig:appendix_roic13_inverse}
\end{figure}

\section{More Visualizations}
\subsection{Visualizations of Line Annotations}
In Sec. 4.7 of the main paper, we study the model sensitivity to different line locations. We provide a group of visualizations in Fig.~\ref{fig:noise_label_vis} to intuitively show the noisy line locations.

\subsection{Visualizations on RoIC13 and Inverse-Text}
More qualitative results on RoIC13 and Inverse-Text are provided in Fig.~\ref{fig:appendix_roic13_inverse}. In the bottom row, some upside-down instances are not detected, which remains a significant challenge when spotting inverse-like instances. Correctly detecting and recognizing inverse-like text requires more discriminative features.

\end{appendices}

\bibliography{sn-bibliography}

\end{document}